\documentclass[lettersize,journal]{IEEEtran}
\usepackage{amsmath,amsfonts}
\usepackage{algorithm}
\usepackage{array}
\usepackage[caption=false,font=normalsize,labelfont=sf,textfont=sf]{subfig}
\usepackage{textcomp}
\usepackage{stfloats}
\usepackage{url}
\usepackage{verbatim}
\usepackage{graphicx}
\usepackage{cite}
\usepackage{makecell}
\usepackage{algpseudocode}
\usepackage{bm}
\usepackage{booktabs}
\usepackage{multirow}
\usepackage{comment}
\usepackage{tikz}
\def\eqref#1{Eq.~(\ref{#1})}

\newcommand{\bb}[1]{{\mathbb{#1}}}
\newcommand*{\Scale}[2][4]{\scalebox{#1}{$#2$}}%

\def\bmzero{{\bm{0}}}

\def\bmmu{{\bm{\mu}}}

\newcommand{\bmI}{\bm{\mathrm{I}}}
\newcommand{\bmX}{\bm{\mathrm{X}}}
\newcommand{\bmXh}{\bm{\mathcal{X}}}

\newcommand{\bmx}{\bm{\mathrm{x}}}

\newcommand{\bbE}{{\mathbb{E}}}

\newcommand{\Var}{\mathrm{Var}}
\newcommand{\Cov}{\mathrm{Cov}}

\def\rmo{{\mathrm{co}}}

\def\rvepsilon{{\bm{\epsilon}}}

\def\gN{{\mathcal{N}}}

\newcommand{\bmJ}{\bm{\mathrm{J}}}
\usepackage{tcolorbox}

\usepackage{xcolor}

\newcommand{\changes}[1]{{\color{black}#1}}
\newcommand{\changess}[1]{{\color{black}#1}}

\begin{document}

\title{Heteroscedastic Diffusion for Multi-Agent Trajectory Modeling}

\author{Guillem Capellera, Antonio Rubio, Luis Ferraz, Antonio Agudo, {\em Member, IEEE}
\thanks{G. Capellera and A. Agudo are with the Institut de Robòtica i Informàtica Industrial, CSIC-UPC, Barcelona, 08028, Spain ({name.surname}@upc.edu).}%
\thanks{G. Capellera, A. Rubio, and L. Ferraz are with Kognia Sports Intelligence, 08960, Sant Just Desvern, Spain ({name.surname}@kogniasports.com).}}

\markboth{IEEE TRANSACTIONS ON PATTERN ANALYSIS AND MACHINE INTELLIGENCE}%
{Capellera \MakeLowercase{\textit{et al.}}: Heteroscedastic Diffusion for Multi-Agent Trajectory Modeling}


\maketitle

\begin{abstract}
Multi-agent trajectory modeling traditionally focuses on forecasting, often neglecting more general tasks like trajectory completion, which is essential for real-world applications such as correcting tracking data. Existing methods also generally predict agents' states without offering any state-wise measure of heteroscedastic uncertainty. Moreover, popular multi-modal sampling methods lack error probability estimates for each generated scene under the same prior observations, which makes it difficult to rank the predictions at inference time. We introduce U2Diffine, a \textbf{unified} diffusion model built to perform trajectory completion while simultaneously offering state-wise heteroscedastic \textbf{uncertainty} estimates. This is achieved by augmenting the standard denoising loss with the negative log-likelihood of the predicted noise, and then propagating the latent space uncertainty to the real state space using a first-order Taylor approximation. We also propose U2Diff, a faster baseline that avoids gradient computation during sampling. This approach significantly increases inference speed, making it as efficient as a standard generative-only diffusion model. For post-processing, we integrate a Rank Neural Network (RankNN) that enables error probability estimation for each generated mode, demonstrating strong correlation with ground truth errors. Our method outperforms state-of-the-art solutions in both trajectory completion and forecasting across four challenging sports datasets (NBA, Basketball-U, Football-U, Soccer-U), underscoring the effectiveness of our uncertainty and error probability estimation.
\end{abstract}


\section{Introduction}
\IEEEPARstart{M}{odeling} trajectories in multi-agent settings is crucial for capturing stochastic human behaviors in various domains, including pedestrian motion prediction~\cite{alahi2016social,gupta2018social, amirian2019social, kosaraju2019social,salzmann2020trajectron++,ngiam2021scene,girgis2021latent, navarro2022social, saadatnejad2023social, xu2023eqmotion}, human pose estimation~\cite{fragkiadaki2015recurrent,jain2016structural,martinez2017human,mao2019learning,mao2020history,aksan2021spatio,cai2020learning,guo2023back}, and sports analytics~\cite{zheng2016generating, zhan2018generating, alcorn2021baller2vec++, hu2022entry, capellera2024footbots, peraltemporally, mao2023leapfrog, xu2023uncovering}. 

Multi-modal generative approaches primarily focus on forecasting future states based on past trajectories. These methods often utilize models based on Generative Adversarial Networks (GANs)~\cite{dendorfer2021mg, gupta2018social}, Conditional Variational Auto-Encoders (CVAEs)~\cite{yuan2021agentformer,xu2022groupnet}, and Denoising Diffusion Probabilistic Models (DDPM)~\cite{ho2020denoising}. DDPMs, in particular, have shown promising results in trajectory forecasting for pedestrian and sports modeling~\cite{mao2023leapfrog, gu2022stochastic}. However, a key limitation of the proposed approaches is their focus on agent-wise metrics, which often neglects scene-level dynamics that are crucial for multi-agent contexts. Additionally, these methods typically require fixed temporal window dimensions, which restricts their adaptability across diverse task settings and scenarios.

The task of \textit{trajectory completion} has emerged as a key advancement. Unlike traditional forecasting, trajectory completion uses both past and/or future observations to infer missing portions of a trajectory~\cite{xu2023uncovering, liu2019naomi, qi2020imitative}. It can also be used to predict the motions of completely unobserved agents based on the movements of surrounding observed ones~\cite{kim2023ball, xu2025sportstraj, capellera2024transportmer}. This capability is especially relevant in sports, where complex multi-agent interactions require models to accurately capture both individual and coordinated team tactical behaviors within fixed spatial coordinates, e.g., a court or field.

Despite these advancements, a critical gap remains: current state-of-the-art rarely provide an estimate of uncertainty for each predicted state. This limitation highlights the need for a state-wise heteroscedastic uncertainty estimation approach to quantify the likelihood of each predicted state. The absence of this measure also presents a further challenge in developing methods to extract a scene-level uncertainty or error probability capable of ranking the reliability of multiple generated modes under the same prior observations.

\begin{figure*}[!t]
\centering
\includegraphics[width=1.0\textwidth, trim={0.5cm 0.8cm 0.5cm 0}]{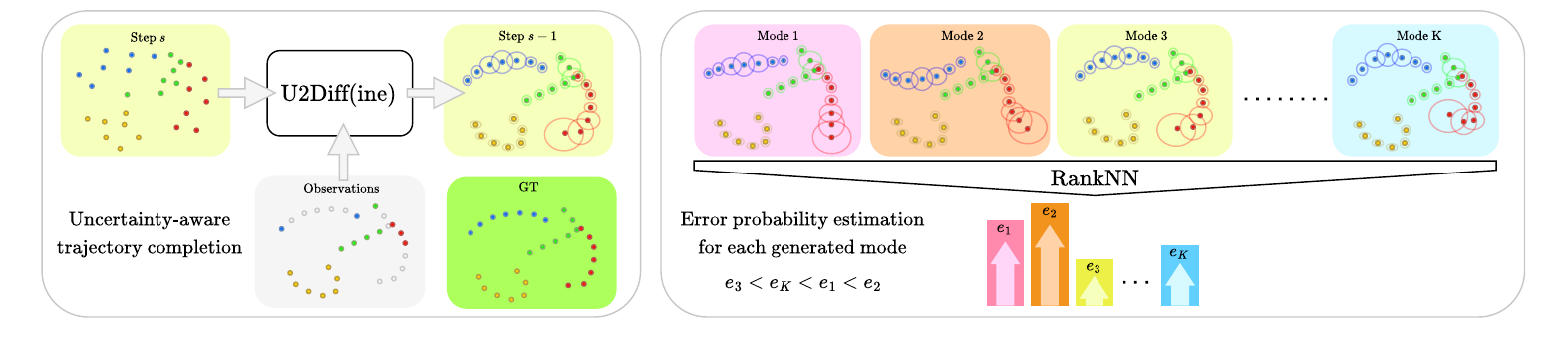}
\caption{\textbf{Uncertainty-aware, unified and interpretable approach for trajectory modeling in multi-agent scenarios.} \textbf{U2Diffine/U2Diff} is a diffusion-based model capable of performing trajectory completion tasks such as forecasting, imputation or inferring totally unseen agents, while also jointly estimating state-wise uncertainty. \textbf{RankNN} is a post-processing operation that infers an error probability for each generated mode under the same prior observations, which is strongly correlated with the error related to the ground truth.}
\label{fig:teaser}
\vspace{-4mm}
\end{figure*}

An early version of this research was presented in~\cite{capellera2025unified}, where we introduced an \textbf{U}nified \textbf{U}ncertainty-aware \textbf{Diff}usion (U2Diff) model. U2Diff was designed to address the general task of multi-agent trajectory completion while predicting per-state heteroscedastic uncertainty by proposing a variance propagation approximation from the latent to the real space (see Fig.~\ref{fig:teaser}). In our initial work, scene-level uncertainty was estimated by averaging the variances of each predicted agent's state, which demonstrated a moderate correlation with scene-level error and provided an unsupervised measure of confidence. To enhance model interpretability, we also introduced a supervised Rank Neural Network (RankNN) to order the generated modes by their proximity to ground truth, yielding error probabilities and achieving high median Spearman correlations between $0.58$ and $0.78$. 

Building upon this foundation, the present work introduces \textbf{U2Diffine}, an extended framework with significant advancements. \changes{We transcend the heuristic limitations of our preliminary work by formally deriving \textbf{Reverse Gaussian Sampling} through a first-order Taylor approximation. We identify that U2Diff corresponds to a simplified special case that relies on a zero-gradient assumption and requires variance delay to maintain accurate variance estimates. U2Diffine, in contrast, eliminates this assumption, providing a \textit{fine}r analytical approach to propagate uncertainty that yields demonstrably superior likelihood estimation}. To further enhance the model's capability to infer rotated uncertainty ellipses, we have coupled this with a bi-variate Gaussian formulation to model the joint $xy$ probability. Our empirical validation rigorously benchmarks the likelihood performance of our proposal against the original U2Diff, albeit with a computational cost approximately four times higher. We further extend our evaluation to trajectory completion and forecasting tasks, showing consistent gains over the state-of-the-art in scene-level accuracy metrics.


We validate the effectiveness of our overall approach using four real-world sports datasets: two of basketball, one of football, and another of soccer; demonstrating substantial improvements over competing methods in scene-level metrics for forecasting and trajectory completion tasks. This work contributes a novel uncertainty-aware approach to trajectory modeling that enhances the reliability of generated trajectories in complex interactive environments like sports. The key contributions are summarized as follows:
\begin{itemize}
    \item We propose a diffusion-based approach for multi-agent trajectory completion, achieving state-of-the-art results.
    \item We augment the diffusion framework with a bi-variate negative log-likelihood loss to enable heteroscedastic uncertainty estimation for each predicted latent state. 
    \item We develop a first-order Taylor approximation for Reverse Gaussian Sampling, propagating latent uncertainty to the output space for precise, robust, and consistent state-wise uncertainty estimates.
    \item We devise a post-processing supervised architecture (RankNN) that provides error probability estimates for each generated mode under a shared prior, enabling a high-correlation with ground truth error.
\end{itemize}

\noindent \changess{\textbf{Practical guidance.} U2Diffine is preferred when calibrated uncertainty is required, as its Taylor propagation and bi-variate modeling yield superior likelihood. U2Diff is preferred when inference latency matters, offering ${\sim}4\times$ faster sampling at comparable displacement error.}

\section{Related Work}
We next review the most relevant work on trajectory modeling, diffusion models, uncertainty estimation, and error probability estimation for each generated mode.

\noindent \textbf{Trajectory Modeling.} Multi-agent trajectory forecasting has advanced significantly by employing various deep generative models to capture human behaviors. Early approaches used Variational Recurrent Neural Networks (VRNNs) focused on modeling the stochasticity inherent in long-term motion~\cite{felsen2018will, zhan2018generating, zheng2016generating, sun2019stochastic, yeh2019diverse}. The field then evolved with GANs~\cite{dendorfer2021mg, gupta2018social,fang2020tpnet, hu2020collaborative, sadeghian2019sophie} and CVAEs~\cite{yuan2021agentformer, mangalam2020not, salzmann2020trajectron++, xu2022groupnet, lee2022muse}, which enabled more diverse and realistic future predictions. More recently, diffusion models have shown great promise in generating diverse plausible future trajectories~\cite{mao2023leapfrog, jiang2023motiondiffuser, gu2022stochastic, rempe2023trace, bae2024singulartrajectory, li2023bcdiff, yang2024diffusion, fu2025moflow}, often outperforming earlier methods in forecasting tasks. Transformer architectures have also pushed the boundaries of trajectory prediction by incorporating explicit hypergraph modeling in their encoder~\cite{lee2024mart} or by integrating visual information, such as the human past poses, as input~\cite{saadatnejad2023social, gao2024multi}. Despite these advancements, many of these prior architectures often suffer from limitations. Their parameters can depend on the temporal horizon, and performance is typically evaluated using individual agent-level metrics, which often overlooks overall scene-level performance. Separately, methods like Graph Variational Neural Networks (GVRNNs)~\cite{xu2023uncovering, omidshafiei2022multiagent} and non-autoregressive techniques~\cite{liu2019naomi,xu2025sportstraj}, have been developed to solve the general trajectory completion task. Building on these foundations, our work introduces a diffusion-based architecture that also integrates forecasting and completion in a unified framework, adaptable to multi-agent scenarios without predefined agent or time dimension constraints. Furthermore, we demonstrate that our method outperforms previous forecasting works on scene-level metrics. 

\noindent \textbf{Unified diffusion models.} Diffusion models have also emerged as powerful generative frameworks for time series~\cite{tashiro2021csdi, alcaraz2022diffusion}. Models like CSDI (Conditional Score-based Diffusion Imputation)~\cite{tashiro2021csdi} have applied score-based diffusion to multi-variate time series imputation as a viable alternative to Gaussian processes for probabilistic modeling.
Our U2Diff/U2Diffine architecture is inspired by CSDI, which we adapted for multi-agent 2D trajectory modeling. We have replaced the traditional Transformer Encoder~\cite{vaswani2017attention} with a bidirectional MambaSSM~\cite{gu2023mamba} to better process temporal dynamics. This change not only improves performance, but also eliminates the need for temporal positional encoding, due to MambaSSM's inherent sequential processing.

\noindent \textbf{Uncertainty estimation.} Despite progress in generative trajectory modeling, state-wise heteroscedastic uncertainty estimation remains an underexplored area. In the autonomous driving domain, some works have focused on estimating uncertainty arising from agent interactions~\cite{tang2023collaborative}, uncertainty in human pose forecasting~\cite{saadatnejad2024toward}, and uncertainty modeling in traffic scenarios~\cite{manas2025uncertainty}. In contrast, in non-ego-centric domains such as pedestrian and sports settings, which are the focus of our work, a diffusion-based approach involves computing global agent-level uncertainty from the past observations~\cite{mao2023leapfrog} to accelerate the sampling process. However, such methods focus solely on forecasting and are not equipped to handle the more general task of trajectory completion. To address uncertainty this broader task, models like CSDI~\cite{tashiro2021csdi} sample multiple modes to approximate a distribution, effectively mimicking a Gaussian process. However, this approximation can be prohibitive in settings of high number of agents or long time horizons. In the image domain, a recent diffusion-based method~\cite{kou2023bayesdiff} introduced a novel uncertainty estimation strategy by applying a Bayesian Laplace approximation to the final layer of a pre-trained score model, enabling pixel-wise variance estimation in latent space. The resulting noise variance is then propagated to the output space through a reverse Gaussian process, with Monte Carlo sampling used to estimate the full covariance. Inspired by the idea of variance propagation, our method extends uncertainty estimation to multi-agent trajectories. We directly compute state-wise heteroscedastic uncertainty by modeling the variance of the predicted noise in the loss function, allowing uncertainty estimation for each agent and timestep. This is followed by a first-order Taylor approximation in the Reverse Gaussian Sampling to propagate uncertainty from latent to real space, avoiding costly Monte Carlo sampling at the expense of gradient computation. To further improve efficiency, we also propose a faster variant, U2Diff, which skips the gradient step while still offering competitive performance.

\noindent \textbf{Probability estimation.} In multi-modal trajectory generation, estimating probabilities for each mode remains relatively unexplored. Existing approaches can be categorized by their scope and modeling assumptions. Ego-agent focused methods assign probabilities using predefined trajectory anchors~\cite{chai2019multipath, phan2020covernet, shi2023trajectory} or by post-processing the predicted trajectory~\cite{zhao2021tnt}. \changes{Agent-level forecasting techniques model probabilities by estimating intent via learnable motion query pairs~\cite{shi2022motion} or use flow-matching frameworks with ranking and classification losses to generate correlated modes~\cite{fu2025moflow,yan2025trajflow}.} Scene-level models~\cite{girgis2021latent, salzmann2020trajectron++} extend probability estimation to the joint multi-agent level, but typically output a fixed number of correlated joint future modes. \changes{A key limitation across these methods is their limited suitability for generative models that produce independent and identically distributed (i.i.d.) scene modes, like ours. Specifically, there is no established mechanism to rank modes by estimating an error probability for each independently sampled mode within such a variable set. To address this, we propose RankNN, a permutation-equivariant post-processing network that estimates error probabilities for each generated scene mode based on agents' trajectories and state uncertainties.}



\section{Revisiting Diffusion Models}
\label{rection_revisiting}

We review DDPM~\cite{ho2020denoising}, which form the foundation of our approach. This generative operates through two stages: a fixed {\em forward diffusion} process, and a learned {\em reverse denoising} one. The forward process systematically corrupts a data sample with Gaussian noise over $S$ steps. The reverse process then learns to iteratively reverse this corruption, starting from pure noise, to generate a new sample from data distribution.

Let $\bmX_0 \sim q(\bmX_0)$ be a sample from the true data distribution. The forward process is a Markov chain that progressively adds Gaussian noise according to a monotonically increasing variance schedule $\{\beta_s\in (0,1)\}^{S}_{s=1}$. At each step $s$, the transition is defined as:
\begin{align}
     q(\bmX_s \mid \bmX_{s-1}) &= \gN\left(\bmX_s; \sqrt{1-\beta_s} \bmX_{s-1}, \beta_s \bmI\right),\\
     q(\bmX_{1:S} \mid \bmX_0) &= \prod_{s=1}^{S} q(\bmX_s \mid \bmX_{s-1}),
    \label{eq:forward}
\end{align}
where $\bmI$ denotes an identity matrix, i.e., the distribution is always represented by a diagonal matrix of variances. The schedule $\{ \beta_s\}$ is designed such that for a sufficiently large $S$, $\bmX_S$ converges to an isotropic Gaussian distribution, i.e., $\bmX_S \sim \gN(\bmzero,\bmI)$.

A key property of this process is that we can sample $\bmX_s$ at any arbitrary step s conditioned on $\bmX_0$ in a closed form. Letting $\alpha_s=1-\beta_s$ and $\hat{\alpha}_s = \prod_{i=1}^s \alpha_i$, the conditional distribution is given by: 
\begin{equation}
q(\bmX_s \mid \bmX_0) = \gN (\bmX_s; \sqrt{\hat{\alpha}_s}\bmX_0, (1-\hat{\alpha}_s)\bmI).
\end{equation}

This allows us to express any noisy sample $\bmX_s$ using the reparameterization trick:
\begin{equation}
\label{eq:sampling_way}
    \bmX_s = \sqrt{\hat{\alpha}_s}\bmX_0 + \sqrt{1-\hat{\alpha}_s} \rvepsilon_s, 
\end{equation}
where  $\rvepsilon_s \sim \gN(\bmzero,\bmI)$. This writes $\bmX_s$ as a linear combination of the initial data sample and a standard Gaussian noise vector, with signal-to-noise ratio controlled by $\hat{\alpha}_s$.

The reverse process generates data by starting with a sample from the prior, $\bmX_S \sim \mathcal{N}(\bmzero, \bmI)$, and iteratively sampling from the reverse conditional distributions $p(\bmX_{s-1} \mid \bmX_s)$. However, the true posterior $q(\bmX_{s-1} \mid \bmX_s)$ is intractable, as it depends on the entire data distribution. Therefore, we approximate it with a neural network $p_{\theta}(\cdot)$ parametrized by $\theta$ as:
\begin{align}
     \label{eq:prior}
     p_\theta(\bmX_{s-1} \mid \bmX_s) &=   \gN(\bmX_{s-1}; \bmmu_\theta(\bmX_s,s),\sigma_\theta(\bmX_s,s)^2\bmI),\\
      p_\theta(\bmX_{0:S}) &= p(\bmX_S) \prod_{s=1}^{S} p_\theta(\bmX_{s-1} \mid \bmX_s), \label{eq:reverse1}
\end{align}
where $p(\bmX_S) = \gN(\bmX_S; \bmzero,\bmI)$ and $\bmmu_\theta(\cdot)$ and $\sigma_\theta(\cdot)^2\bmI$ denote the predicted mean and diagonal covariance, respectively, for each transition in the reverse chain.

The mean in~\eqref{eq:prior} is commonly expressed in terms of the predicted noise $\rvepsilon_\theta (\bmX_s,s)$ at step $s$, as:
\begin{align}
  \bmmu_\theta(\bmX_s,s)&=\frac{1}{\sqrt{\alpha_s}} \left(\bmX_s-\frac{\beta_s}{\sqrt{1-\hat{\alpha}_s}} \rvepsilon_\theta (\bmX_s,s) \right),  \label{eq:mu} 
\end{align}
where $\rvepsilon_\theta$ is a trainable denoising function. For brevity, we will consider $\rvepsilon_\theta (\bmX_s) := \rvepsilon_\theta (\bmX_s,s)$.

In general, to infer the covariance matrix, the variance is assumed to be $\sigma_\theta(\bmX_s,s)^2 = \frac{1 - \hat{\alpha}_{s-1}}{1 - \hat{\alpha}_s} \beta_s$, i.e., it does not depend on the predicted noise. This parametrization is equivalent to rescaled score in score-based generative models~\cite{song2020score}. Then, the reverse process objective is simplified by minimizing the following function:
\begin{equation}
  \mathcal{L}_{\mathrm{simple}} = 
  \bb{E}_{\bmX_0 , \rvepsilon ,s}
  || \rvepsilon_s - \rvepsilon_{\theta}(\bmX_s) ||_2^2 ,
  \label{eq:loss}
\end{equation}
where $\rvepsilon_s$ is a random but known noise. Nichol {\em et al.}~\cite{nichol2021improved} found that learning the variance explicitly $\sigma_\theta(\bmX_s,s)^2$, and therefore directly minimizing the Variational Lower Bound (VLB), improved the likelihood in images.

\section{Problem Statement: Multi-Agent Trajectory Modeling}
Let us consider a set of $N \in \mathbb{N}$ agent observations in a given time instant $t$, denoted as $\bmx_t = \{\bmx^n_t\}$ with $n=\{1,\ldots,N\}$, where each observation contains the $(x,y)$ locations. We can now collect $T$ observations along time for every agent, defining a scene tensor $\bmX$ where all $\bmx^{n}_{t}$ with $t=\{1,\ldots,T\}$ are considered. Trajectory completion aims at inferring missing or unobserved entries of a data structure based on the visible ones. Given a tensor of partial observations defined as $\bmX^\rmo$ and a $T \times N$ binary conditioning mask $\mathbf{M}$ to encode by 1 the visible observations and by 0 the unobserved ones, the goal is to find a function $f(\cdot)$ to infer the full observations such that:
\begin{equation}
\label{completion}
   \bmX=f(\bmX^\rmo, \mathbf{M} ).
\end{equation}

Particularly, in this paper we propose to model multi-agent trajectories by leveraging per-observation uncertainty estimation as:
\begin{equation}
\label{completion_distrib}
\Scale[0.95]{
   p(\bmX \mid \bmX^\rmo, \mathbf{M}) = \gN(\bmX; f^\mu(\bmX^\rmo, \mathbf{M}), f^{\Sigma}(\bmX^\rmo, \mathbf{M})),}
\end{equation}
where $f^\mu(\cdot)$ and $f^{\Sigma}(\cdot)$ denote the function to extract mean and covariance matrix, respectively. As we propose to employ a generative model to handle the previous problem, at inference time the method obtains $K\in \mathbb{N}$ modes or scenes according to the same prior observations such that:
\begin{equation}
\label{inference}
   p(\bmX^k \mid \bmX^\rmo, \mathbf{M}) \quad \forall k \in \{1, \ldots, K\}.
\end{equation}

Once the trajectory completion problem is addressed, we propose estimating an error probability for each mode, \changes{to rank them by deviation from the ground-truth trajectories.}


\section{Unified Uncertainty-aware Diffusion}

In this section we describe how to exploit probabilistic diffusion models to address trajectory completion in multi-agent scenarios. Our work is inspired by CSDI~\cite{tashiro2021csdi} that used a diffusion probabilistic framework for handling one-dimensional multivariate time-series imputation, exploiting visible observations to infer the non-visible ones. Following their approach, our denoiser function ingests visible observations $\bmX^\rmo$. For notational brevity, we will write the denoiser as $\rvepsilon_{\theta}(\bmX_s) = \rvepsilon_{\theta}(\bmX_s, s, \bmX^\rmo)$.

Our Unified Uncertainty-aware Diffusion approach to infer the set of distributions was introduced in~\eqref{inference}. This method specifically targets the generation and reconstruction of agents states $(x,y)$, while simultaneously quantifying their data-dependent uncertainty. We achieve this by explicitly modeling the uncertainty in the reverse noise estimation. Rather than a single estimate, we learn a full probability distribution over the noise vector for each agent at each frame. This formulation captures data-dependent (aleatoric) heteroscedastic uncertainty, which intrinsically varies across agents, temporal frames, and spatial dimensions. This distinguishes our approach from epistemic uncertainty modeled in ELBO frameworks~\cite{ho2020denoising, nichol2021improved}, which originates from the Kullback-Leibler's divergence between the learned reverse process $p_\theta(\bmX_{s-1}|\bmX_s)$ and the true posterior $q(\bmX_{s-1} \mid \bmX_s, \bmX_0)$, a non-data dependent distribution with a scheduled variance $\beta_s$.

Formally, we define the conditional distribution over the reverse noise as a bi-variate Gaussian:
\begin{equation}
 \gN(\rvepsilon_s; \rvepsilon^{\mu}_\theta(\bmX_s), \rvepsilon^{\Sigma}_\theta(\bmX_s)).
\end{equation}
Here $\rvepsilon^{\mu}_\theta$ represents the predicted mean noise-state, and $\rvepsilon^{\Sigma}_\theta$ represents a $2 \times 2$ symmetric positive-definite covariance matrix. This matrix captures spatial correlations between $x$ and $y$ components, enabling the representation of directional uncertainty and rotated uncertainty ellipses. This capability is especially valuable in trajectory modeling, where motion uncertainty often has directional dependencies. A specific case, where the covariance matrix is diagonal, corresponds to a uni-variate model.

To learn these parameters, we introduce a loss term, $\mathcal{L}_\mathrm{NLL}$, which maximizes the likelihood of the true noise $\rvepsilon_s$ under the predicted state-wise noise distribution. This is equivalent to minimizing the Negative Log-Likelihood (NLL):
\begin{equation}
\Scale[1.0]{
\mathcal{L}_{\mathrm{NLL}} = \frac{1}{2}\bbE\left[\log(2\pi \left| \rvepsilon^{\Sigma}_\theta \right|^{\frac{1}{2}}) + \frac{1}{2} \omega^\top (\rvepsilon^{\Sigma}_\theta)^{-1} \omega\right]}, 
\label{eq:loss_nll}
\end{equation}
with $\omega=\rvepsilon_s - \rvepsilon^{\mu}_\theta$. A normalization factor of 2 is included to report the NLL per coordinate. We combine this regularizer with the standard denoising objective (see \eqref{eq:loss}) to obtain the total loss function:
\begin{equation}
  \mathcal{L}_{\mathrm{total}} = \mathcal{L}_{\mathrm{simple}} + \lambda \mathcal{L}_{\mathrm{NLL}},
\end{equation}
where $\lambda$ is a small weight factor (around $0.01$) that balances the influence of $\mathcal{L}_\mathrm{NLL}$ without overwhelming the primary learning objective. Following the approach in~\cite{nichol2021improved}, we apply a stop-gradient to the mean term, $\rvepsilon^{\mu}_{\theta}(\bmX_s)$, so that $\mathcal{L}_\mathrm{NLL}$ focuses exclusively on learning the variance.

\subsection{\changes{Reverse Gaussian Sampling}}
\label{subsec:rev_gaussian}
During sampling, we explicitly model state-dependent uncertainty by treating the reverse process as Reverse Gaussian Sampling. \changess{Throughout, $\Var(\cdot)$ denotes the variance operator; applied to a vector-valued variable (2D in our setting), it returns the $2\times2$ covariance matrix. We use the terms variance and covariance interchangeably to refer to the output of this operator, and $\Cov(\cdot, \cdot)$ denotes the cross-covariance between two variables.} \changes{We define the latent state as a Gaussian random variable $\bmXh_s \sim \gN(\bmX_s, \Var(\bmXh_s))$, where $\bmX_s = \bbE[\bmXh_s]$ is the point estimate (mean) and $\Var(\bmXh_s)$ denotes the associated covariance matrix, which captures the heteroscedastic uncertainty.} Using Denoising Diffusion Implicit Model (DDIM)~\cite{song2020denoising} with skip interval $\zeta$, this state evolves as:

\vspace{-4mm}
\begin{equation}
\label{eq:ddim}
\Scale[0.99]{
  \bmXh_{s-\zeta} = \underbrace{\sqrt{\frac{\hat{\alpha}_{s-\zeta}}{\hat{\alpha}_{s}}}}_{a_s}\bmXh_s + \underbrace{\left(\gamma_{s-\zeta} - \sqrt{\frac{\hat{\alpha}_{s-\zeta}}{\hat{\alpha}_{s}}}\gamma_s \right)}_{b_s} \rvepsilon_\theta(\bmXh_s)},
\end{equation}
where $\gamma_{s} = \sqrt{1 - \hat{\alpha}_s}$ and $\gamma_{s-\zeta} = \sqrt{1 - \hat{\alpha}_{s-\zeta}}$. Under the standard noise schedule, we have $\hat{\alpha}_{s - \zeta} > \hat{\alpha}_s$ \changes{(and consequently $\gamma_{s-\zeta}< \gamma_{s}$)}, implying coefficients $a_s > 1$ and $b_s < 0$.


A direct Monte Carlo estimate of the moments at each step is computationally prohibitive. Instead, we approximate the moments by linearizing the score network $\rvepsilon_\theta$ around the current mean $\bmX_s$ via a first-order Taylor expansion:
\begin{align}
    \rvepsilon_\theta^\mu(\bmXh_s) &=  \rvepsilon_\theta^\mu(\bmX_s) + \bmJ_s(\bmXh_s - \bmX_s )  + \mathcal{O}(||\bmXh_s - \bmX_s||^2) ,\nonumber \\
    \rvepsilon_\theta^\Sigma(\bmXh_s) &=  \rvepsilon_\theta^\Sigma(\bmX_s) + \bmJ_s^\Sigma(\bmXh_s - \bmX_s )  + \mathcal{O}(||\bmXh_s - \bmX_s||^2), \nonumber
\end{align}
\changes{where $\bmJ_s = \nabla_{\bmX} \rvepsilon_\theta^\mu(\bmX)_{\mid \bmX=\bmX_s}$ and $\bmJ_s^{\Sigma} = \nabla_{\bmX} \rvepsilon_\theta^\Sigma(\bmX)_{\mid \bmX=\bmX_s}$ are per-state $2 \times 2$ Jacobian matrices evaluated at the current mean.} 


\noindent\textbf{Mean propagation.} Taking the expectation of \eqref{eq:ddim} and exploiting $\bbE[\bmXh_s - \bmX_s]=\bmzero$:
\begin{align}
    \label{eq:expected_ddim}
    \bbE[\bmXh_{s-\zeta}] = \bmX_{s - \zeta} &= a_s\bmX_s + b_s\bbE[\rvepsilon_\theta(\bmXh_s)] \nonumber \\
    &= a_s\bmX_s + b_s\bbE[\bbE[\rvepsilon_\theta(\bmXh_s) \mid \bmXh_s]] \nonumber \\
    &= a_s\bmX_s + b_s\bbE[\rvepsilon_\theta^\mu(\bmXh_s)] \nonumber \\
    &\approx a_s\bmX_s + b_s\rvepsilon_\theta^\mu(\bmX_s).  
\end{align}

\noindent\textbf{Variance propagation.} Applying the variance operator to \eqref{eq:ddim}:
\begin{align}
    \label{eq:var_decomp}
    \Var(\bmXh_{s-\zeta}) &=  a_{s}^{2}\Var(\bmXh_s) + b_{s}^{2} \Var(\rvepsilon_\theta(\bmXh_s))  \\ &+ a_s b_s [ \Cov(\bmXh_s,\rvepsilon_\theta(\bmXh_s) ) + \Cov(\rvepsilon_\theta(\bmXh_s), \bmXh_s)] \nonumber .
\end{align}
For the noise variance term we use the law of total variance:
\begin{align}
    \Var(\rvepsilon_\theta(\bmXh_s)) &= \bbE[\Var(\rvepsilon_\theta(\bmXh_s) \mid \bmXh_s)] + \Var(\bbE[\rvepsilon_\theta(\bmXh_s) \mid \bmXh_s]) \nonumber \\
    &= \bbE[\rvepsilon^{\Sigma}_{\theta}(\bmXh_s)] + \Var(\rvepsilon^{\mu}_{\theta}(\bmXh_s))  \nonumber \\
    &\approx \rvepsilon^{\Sigma}_{\theta}(\bmX_s) + \bmJ_s \Var(\bmXh_s) \bmJ_s^\top ,
\end{align}
\changes{where the second term follows from $\Var(\bmX_s)=\bmzero$ (since $\bmX_s$ is deterministic) and $\Var(\bmJ_s (\bmXh_s - \bmX_s)) = \bmJ_s \Var(\bmXh_s) \bmJ_s^\top$.}
Similarly, for the \changess{cross-covariance} terms:
\begin{align}
    \Cov(\bmXh_s,\rvepsilon_\theta(\bmXh_s)) &= \bbE[\Cov(\bmXh_s,\rvepsilon_\theta(\bmXh_s) \mid \bmXh_s)] \nonumber \\& \quad + \Cov(\bbE[\bmXh_s \mid \bmXh_s], \bbE[\rvepsilon_\theta(\bmXh_s) \mid \bmXh_s]) \nonumber \\
    &= \bbE[\bmzero] + \Cov(\bmXh_s,\rvepsilon^{\mu}_{\theta}(\bmXh_s)) \nonumber \\
    & \approx \Cov(\bmXh_s,\rvepsilon^{\mu}_{\theta}(\bmX_s) + \bmJ_s (\bmXh_s - \bmX_s))  \nonumber \\
    &= \Cov(\bmXh_s,\bmJ_s \bmXh_s) = \Var(\bmXh_s) \bmJ_s^\top ,
\end{align}
\changes{using $\Cov(\bmXh_s,\rvepsilon^{\mu}_{\theta}(\bmX_s))=\bmzero$ and $\Cov(\bmXh_s,\bmX_s)=\bmzero$.} By symmetry, $\Cov(\rvepsilon_\theta(\bmXh_s),\bmXh_s) \approx \bmJ_s  \Var(\bmXh_s)$. 


\noindent Substituting these into \eqref{eq:var_decomp} yields the update rule:
\begin{equation}
\label{eq:var_ddim}
\begin{split}
\Var(\bmXh_{s-\zeta}) \approx & (a_s \bmI + b_s \bmJ_s) \Var(\bmXh_{s}) (a_s \bmI + b_s \bmJ_s)^\top \\
&+ b_s^2 \rvepsilon^{\Sigma}_{\theta}(\bmX_s).
\end{split}
\end{equation}

\begin{algorithm}[t]
\caption{\changess{Reverse Gaussian Sampling (U2Diffine / U2Diff)}}
\label{alg:rev_gaussian_sampling}
\begin{algorithmic}[1]
\Require $\bmX^\rmo$ (observations), $\rvepsilon_\theta$ (trained denoiser), $\bar{\zeta}$ (skip interval), $\hat{s}$ (variance-delay step).
\Ensure $\bmX_0$, $\Var(\bmXh_0)$.
\State $\bmX_S \sim \gN(\bmzero,\bmI)$, $\Var(\bmXh_S) \leftarrow \bm0$
\State $\mathcal{D} \leftarrow [S, S-\bar{\zeta}, \dots, \bar{\zeta}, 1]$
\For{$i = 0, \dots, |\mathcal{D}|-2$}
    \State $s \leftarrow \mathcal{D}[i]$, $\zeta \leftarrow \mathcal{D}[i] - \mathcal{D}[i+1]$
    \State Forward denoiser: $\rvepsilon_\theta^{\mu}(\bmX_s), \rvepsilon_\theta^{\Sigma}(\bmX_s)$
    \State Estimate $\bmX_{s-\zeta}$ via \eqref{eq:expected_ddim}
    \If{$s > \hat{s}$} \Comment{delay phase; inactive for U2Diffine}
        \State $\Var(\bmXh_{s-\zeta}) \leftarrow \bmzero$
    \Else
        \State $\bmJ_s \leftarrow \begin{cases} \nabla_{\bmX} \rvepsilon_\theta^\mu(\bmX)_{\mid \bmX=\bmX_s} & \text{U2Diffine} \\ \bm0 & \text{U2Diff} \end{cases}$
        \State Estimate $\Var(\bmXh_{s-\zeta})$ via \eqref{eq:var_ddim}
    \EndIf
\EndFor
\State Forward denoiser: $\rvepsilon_\theta^\mu(\bmX_1)$
\State Estimate $\bmX_0$ via \eqref{eq:expected_ddim} \Comment{final denoising step}
\State $\Var(\bmXh_0) \leftarrow \Var(\bmXh_1)$ \Comment{copy variance}
\end{algorithmic}
\end{algorithm}

Governed by the negative coefficient $b_s$ derived from the noise schedule, the term $\bmJ_s$ regularizes and stabilizes variance propagation within the Reverse Gaussian Process, effectively preventing variance inflation.

\noindent \textbf{Algorithm details.} We initialize the variance of $\bmXh_{S}$ to zero, $\Var(\bmXh_{S}) = \mathbf{0}$. This aligns with the approach in Kou {\em et al.}~\cite{kou2023bayesdiff} to represent zero heteroscedastic uncertainty at the initial noise step $S$. For the last denoising step $s=1$, we compute $\bmX_0$ from $\bmX_1$ and assign $\Var(\bmXh_0) = \Var(\bmXh_1)$. The effective denoising steps are defined as $\mathcal{D} = [S, S-\bar{\zeta}, \ldots, \bar{\zeta}, 1]$, where $\bar{\zeta}$ represents the fixed skip interval. You can find more comprehensive details in Algorithm~\ref{alg:rev_gaussian_sampling}. 

In practice, while per-state 2$\times$2 Jacobian matrix, $\bmJ_s$, is computationally tractable  using standard automatic differentiation libraries, its full form can lead to practical challenges. Specifically, the full Jacobian can be ill-conditioned or contain large off-diagonal elements, introducing numerical instabilities that result in non-positive-definite covariance estimates. To ensure robust and positive-definite covariance estimates, we approximate $\bmJ_s$ by its diagonal. Our primary proposed method, \textbf{U2Diffine}, combines this diagonal Jacobian approximation with an expressive bi-variate uncertainty model. 

For applications requiring faster inference, we define an accelerated variant, \textbf{U2Diff}, which assumes $\bmJ_s = \bmzero$ and uses a simpler uni-variate uncertainty model. While this reduces computation time by over 4$\times$, it risks variance inflation, which we mitigate by delaying the start of uncertainty propagation until an intermediate step $\hat{s} < S$. With $S=50$ and $\bar{\zeta}=10$, our experiments found an optimal $\hat{s}$ of $30$ (see implementation details). \changess{This propagation delay applies only to U2Diff variant; for U2Diffine we set $\hat{s} = S$.}

\begin{figure}[t!]
\vspace{-4mm}
  \centering
  \includegraphics[width=0.99\linewidth]{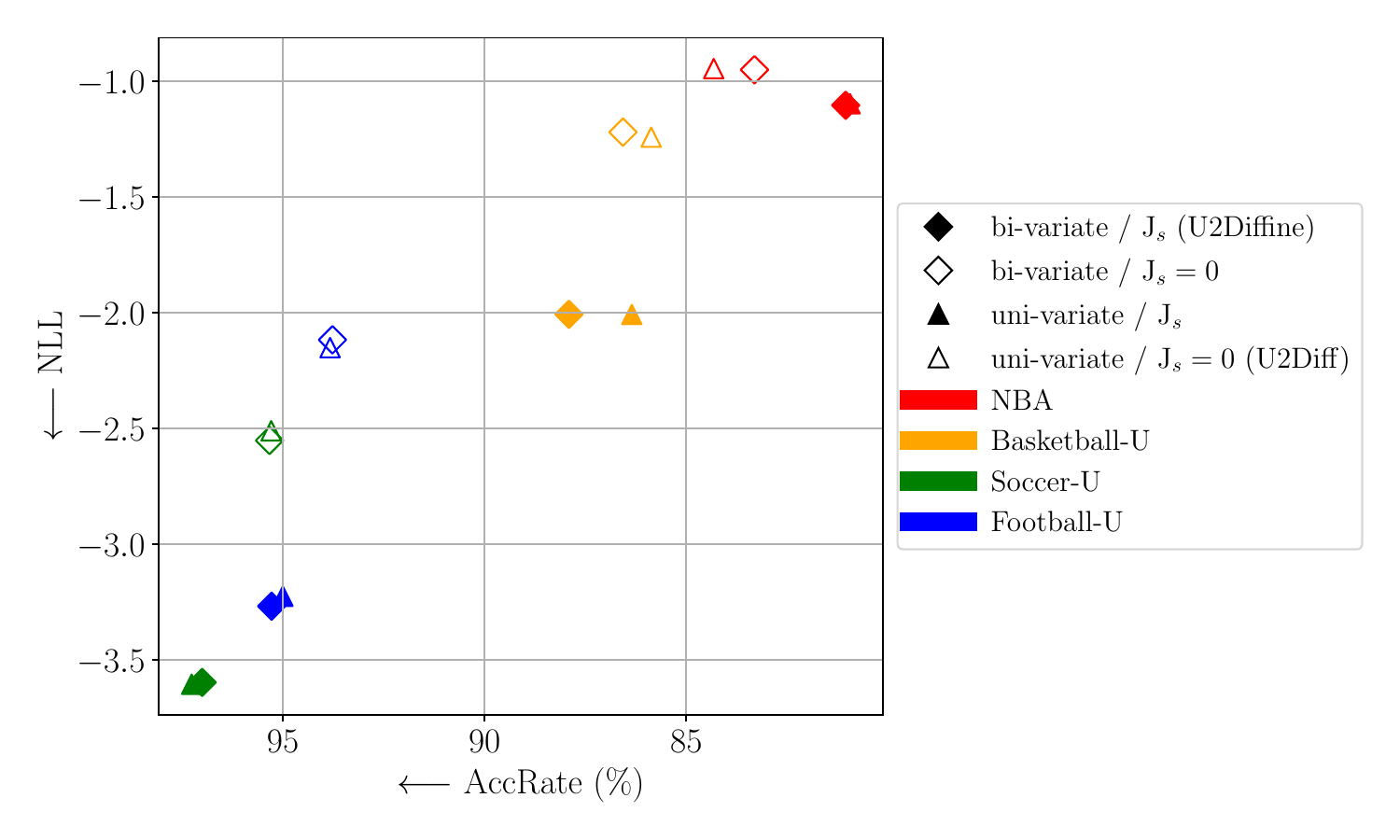}
  \vspace{-8mm}
\caption{\textbf{Reverse Gaussian Sampling}. Accuracy Rate (AccRate) and Negative Log-Likelihood (NLL) results under different configurations of the Reverse Gaussian Sampling procedure.}
  \label{fig:ablation_jacobian}
  \vspace{-3mm}
\end{figure}

Figure~\ref{fig:ablation_jacobian} presents an ablation study investigating Jacobian approximation and uncertainty modeling. For Jacobian approximation, we compare the impact of setting $\bmJ_s = \bmzero$, representing no gradient computation, against using a diagonal Jacobian $\bmJ_s$. For uncertainty modeling, we assess both uni-variate and bi-variate approaches. Performance is evaluated using NLL$\downarrow$ (lower is better), and accuracy rate, AccRate$\uparrow$ (higher is better), with both metrics computed over $K=20$ generated modes. The AccRate quantifies the percentage of ground-truth states falling within the predicted uncertainty ellipse at 95\% confidence (further details are in Section~\ref{sec:ranknn}). Our findings clearly indicate that incorporating Jacobian prediction significantly improves NLL across all datasets. 

\begin{figure*}[t!]
  \centering
  \includegraphics[width=1.00\linewidth,trim={1.5cm 1cm 4cm 0}]{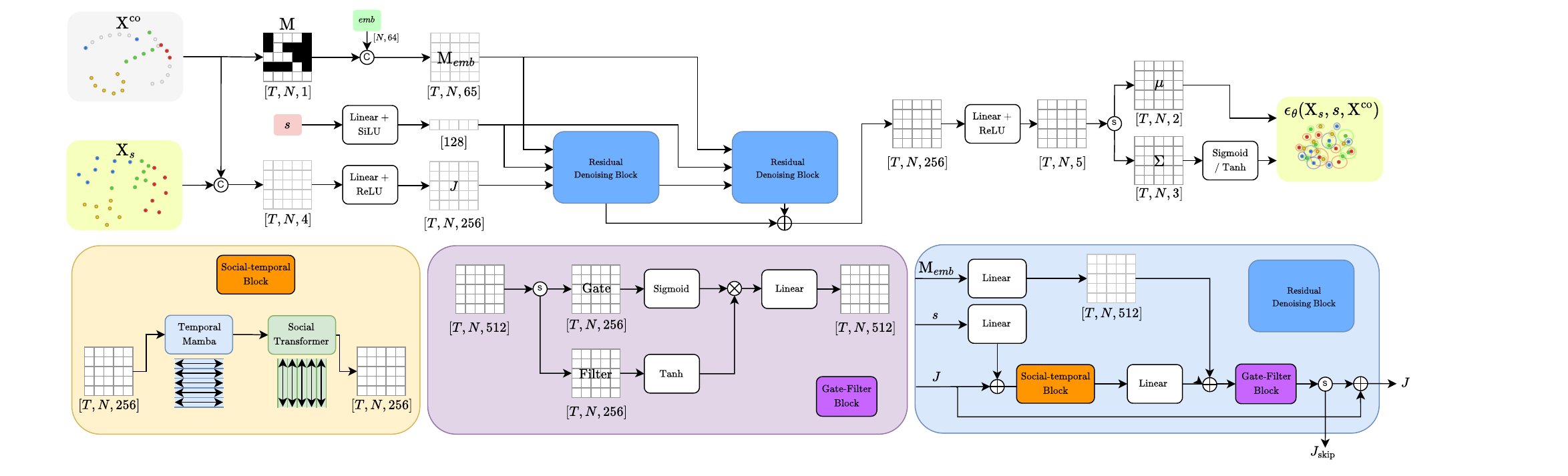}
\caption{\textbf{U2Diffine architecture}. The input consists of noisy trajectories ($\bmX_s$), observations ($\bmX^\rmo$), a mask ($\mathbf{M}$), and the current denoising step ($s$). These inputs are embedded and processed through two Residual Denoising Blocks. Within these blocks, a Social-temporal Block combines a Temporal Mamba and a Social Transformer to effectively model complex temporal and social dynamics. Finally the skip connection outputs, $J_\text{skip}$, are summed and projected to form the mean and the covariance of a bi-variate Gaussian distribution for multi-agent trajectory completion.}
  \label{fig:u2diff_arch}
  \vspace{-4mm}
\end{figure*}

\subsection{Architecture}
We introduce our multi-agent trajectory completion architecture, which is a modified version of the CSDI~\cite{tashiro2021csdi} architecture designed to seamlessly integrate uncertainty estimation. The first modification replaces temporal Transformer Encoder with MambaSSM~\cite{gu2023mamba} for temporal processing, which improves performance during the denoising process. Furthermore, the final linear head of the architecture is redesigned to output both the mean $\rvepsilon^{\mu}_\theta(\bmX_s)$ and the Cholesky parameters of the 2x2 covariance matrix $\rvepsilon^{\Sigma}_\theta(\bmX_s)$, allowing for uncertainty estimation. The full architecture is depicted in Fig.~\ref{fig:u2diff_arch}. We next present the key components of our architecture.


\noindent \textbf{Input embeddings.} Initially, the observed trajectories $\bmX^\rmo$ collected in a tensor with dimensions $T \times N \times 2$, which contain zero values $(0,0)$ for unobserved states, are concatenated with the noised sample $\bmX_s$ of the same dimension, producing a combined tensor with dimensions $T \times N \times 4$. This tensor is then transformed into an embedding tensor $J$ through a linear layer followed by a ReLU activation, resulting in dimensions $T \times N \times 256$. The denoising step index $s$ is also embedded into a 128-dimensional space using a linear layer followed by a SiLU activation function before being incorporated into each residual denoising block. Additionally, a learnable embedding of dimension $64$ is assigned for each agent, referred to as $emb$, and concatenated with the binary mask $\mathbf{M}$, forming the mask/embedding tensor, denoted as $\mathbf{M}_{emb}$. These embeddings are subsequently processed sequentially through two identical residual denoising blocks. 

\noindent \textbf{Residual denoising block.} Each block processes three inputs: the embedding tensor $J$, the embedded denoising step $s$, and the mask/embedding tensor $\mathbf{M}_{emb}$. First, the embedding of $s$ is projected to a 256-dimensional space using a linear transformation to align with the dimensionality of $J$. This transformed embedding is then summed to each state in $J$. The resulting tensor passes through the \textbf{Social-temporal Block} (see Fig.~\ref{fig:u2diff_arch}), which comprises two main components. First, temporal processing is performed independently for each agent using a bidirectional version of the original MambaSSM~\cite{gu2023mamba}, which we term \textit{Temporal Mamba}. In this step, we compute two separate embeddings for each agent coming from the forward and reverse pass through the MambaSSM~\cite{gu2023mamba}. These two embeddings are then summed to capture both past and future temporal information. Second, social processing is conducted to capture interactions between agents at each timestep using a Transformer Encoder~\cite{vaswani2017attention, devlin2019bert}, termed the \textit{Social Transformer}. This decoupling of temporal and social processing allows encoding scenes with variable timesteps and agents without fixing the temporal and social dimensions. 

Within each residual block, the binary mask $\mathbf{M}_{emb}$ is projected into $J$ to encode agent information while indicating which states are observed and unobserved. The resulting tensor undergoes through a \textbf{Gate-Filter Block} which outputs a 512-dimensional tensor which is then split in two of 256-dimensions: the first is a refined version of $J$ that is passed to the next residual denoising block. The second tensor serves as the skip connection output, $J_\text{skip}$, which is stored from each of the two blocks for later use.

\noindent \textbf{Output tensor.} The $J_\text{skip}$ outputs from all residual denoising blocks are aggregated via summation. These aggregated features are then processed by a linear layer and a ReLU activation, producing a $T \times N \times 5$ tensor. This tensor fully parameterizes a bi-variate Gaussian distribution for the predicted noise, $\rvepsilon_\theta$. Conceptually, it is decomposed into the mean vector $\rvepsilon^{\mu}_\theta(\bmX_s, s, \bmX^\rmo)$, a $T \times N \times 2$ tensor, and the parameters governing the covariance matrix, $\rvepsilon^{\Sigma}_\theta(\bmX_s, s, \bmX^\rmo)$. The covariance is parameterized by its Cholesky decomposition components: two standard deviation terms (output as $T \times N \times 2$ tensor) and a correlation term (output as a $T \times N \times 1$ tensor). The standard deviation outputs are passed through a Sigmoid activation to constrain them to $(0,1)$, while the correlation term is constrained to $(-1,1)$ via a Tanh activation. These two output streams are illustrated as $\mu$ and $\Sigma$ in Fig.~\ref{fig:u2diff_arch}. For the U2Diff variant, the correlation term is set to zero.

\begin{figure*}[t!]
  \centering
  \includegraphics[width=0.99\linewidth,trim={1.5cm 0.8cm 0.8cm 0}]{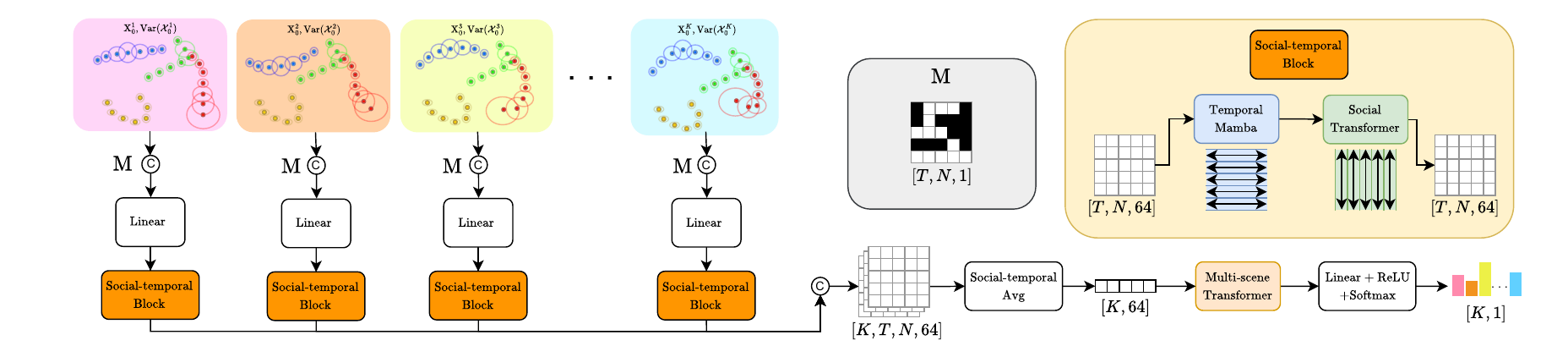}
\caption{\textbf{RankNN architecture}. The model's input for each generated scene consists of the mean ($\bmX_0$), the square roots of the covariance ($\Var(\bmXh_0)$) eigenvalues, and the binary mask ($\mathbf{M}$). These are embedded and processed through a Social-temporal Block. After that, it creates scene-level embeddings by averaging over time and agents. These embeddings are then fed into a Multi-scene Transformer to account for dependencies between different scenes, and the final result is used to predict error probabilities.}
  \label{fig:ranknn_architecture}
  \vspace{-4mm}
\end{figure*}

\section{Rank Neural Network}
\label{sec:ranknn}
\changess{To quantify scene-level uncertainty, we introduce AvgUcty, the mean semi-axis length of the predicted uncertainty ellipses, computed from the eigenvalues $\lambda_1$, $\lambda_2$ of $\Var(\bmXh_0)$ as $\frac{1}{2}(\sqrt{\lambda_1} + \sqrt{\lambda_2})$ averaged over all states in the scene.}

For a given scene prior observations and its $K$ generated modes, ideally the set of AvgUcty per-mode values and the set of corresponding their error values would correlate positively. In other words, higher AvgUcty values correlate with higher error values. The chosen scene-level metric for the error is the Scene Average Displacement Error (SADE), defined as:
\begin{equation}
\label{eq:sade}
    \text{SADE} =\frac{\sum_{n=1}^{N} \sum_{t=1}^{T} \left\lVert \mathbf{\hat{x}}^{n}_{t} - \mathbf{x}^{n}_{t} \right\rVert_2 (1 - \mathbf{m}_{t}^{n})}{\sum_{n=1}^{N} \sum_{t=1}^{T} (1 - \mathbf{m}_{t}^{n})}  \,,
\end{equation}
where $\mathbf{\hat{x}}^{n}_{t}$ and $\mathbf{x}^{n}_{t}$ are the estimation and the corresponding ground truth, respectively, and $\mathbf{m}^{n}_t$ is the value of $\mathbf{M}$ indicating if the $n$-th agent at timestep $t$ is observed or not. 

While AvgUcty provides a straightforward estimation of scene-level uncertainty, it may not fully capture its correlation with SADE. To address this, we introduce a novel learning-based approach that assigns an error probability score, $e$, to each mode, summing to 1 across the $K$ modes. These error probabilities are expected to show a stronger correlation with SADE values compared to AvgUcty. Specifically, we propose a RankNN model, which takes the $K$ generated modes, along with their predicted means and variances, and outputs logits that align with the SADE. 

The objective function to maximize is the Spearman correlation coefficient ($\rho$) between the SADE values and the estimated $e$ ones across all $K$ modes. This coefficient evaluates the monotonic relationship between these two sets. Let $e^k$ and SADE$^k$ represent the error probability estimation and the SADE, respectively, for the mode $k\in \{1, \dots, K\}$. This coefficient, defined as the Pearson correlation between rank variables, is computed by first converting each pair ($e^k$, SADE$^k$) for all $K$ modes into differentiable ranks, denoted ($R$[$e^k$], $R$[SADE$^k$]), with $R$[·] being the differentiable rank operator~\cite{blondel2020fast}. Therefore we can express $\rho$ as:
\begin{equation}
\small{
\label{eq:spearman}
 \rho = \frac{1}{K} \sum_{k=1}^K \left( \frac{(R[e^k] - \overline{R[e]}) \cdot (R[\text{SADE}^k] - \overline{R[\text{SADE}}])}{\| R[e^k] - \overline{R[e]} \| \cdot \| R[\text{SADE}^k] - \overline{R[\text{SADE}]} \|} \right), }
\end{equation}
where the terms $\overline{R[e]}$ and $\overline{R[\text{SADE}}]$ are the mean values over the $K$ generated modes. The normalization in the denominator ensures that this correlation is bounded within $(-1, 1)$. 

\subsection{Architecture}
Figure~\ref{fig:ranknn_architecture} shows the complete RankNN architecture. The input consists of the mean $\bmX_0$ concatenated with the square roots of the eigenvalues from the covariance matrix $\Var(\bmXh_0)$ for each state. This forms a $K \times T \times N \times 4$ tensor, which is then expanded into a $K \times T \times N \times 5$ tensor by adding the binary mask $\mathbf{M}$ repeated $K$ times.

The input tensor is embedded into a 64-dimensional space before being processed by a Social-temporal Block. Within this block, a Temporal Mamba captures individual agent dynamics, with operations repeated across $K \times N$ elements. Subsequently, a Social Transformer models social interactions for each timestep, performing operations across $K \times T$. The resulting tensor with dimensions $K \times T \times N \times 64$ is then averaged across the time ($T$) and agents ($N$) axes. This yields a scene-level embedding tensor of dimensions $K \times 64$. This tensor is then fed into a Transformer Encoder to perform attention across the $K$ scenes, enabling an efficient ranking. \changes{We refer to this operation as the \textit{Multi-scene Transformer}, which is permutation-equivariant with respect to the scene ordering, as no positional encoding is applied along the mode dimension $K$}. Finally, a linear layer with ReLU activation generates a vector of length $K$. This vector is then normalized using a softmax function to produce the error probabilities $\{e^1, \dots, e^K\}$. Notably, similar to our U2Diff/U2Diffine architectures, this setup offers flexibility, as it does not require a fixed number of timesteps, agents, or generated modes.

\section{Experimental design}
\label{sec:experimental_design}

\subsection{Datasets}
For trajectory completion, we evaluate on three team sports datasets~\cite{xu2025sportstraj}: Basketball-U, Football-U, and Soccer-U. \textbf{Basketball-U} derives from NBA dataset~\cite{zhan2018generating} with 93,490 training and 11,543 testing sequences, each spanning 50 frames (8 seconds) capturing $(x,y)$ coordinates for 10 players and the ball. \textbf{Football-U}, based on the NFL Big-Data-Bowl\footnote{https://github.com/nfl-football-ops/Big-Data-Bowl} dataset, contains 10,762 training and 2,624 testing sequences of 50 frames, tracking $(x,y)$ locations for 22 players and the ball. \textbf{Soccer-U}, built from SoccerTrack\footnote{https://github.com/AtomScott/SportsLabKit} dataset, provides 9,882 training and 2,448 testing sequences, each also 50 frames, recording $(x,y)$ positions for 22 players and the ball. In~\cite{xu2025sportstraj}, five masking strategies are defined for trajectory completion, including forecasting futures, imputing in-between states, and inferring the state of over five unobserved agents.

For trajectory forecasting, we use the NBA SportVU dataset (\textbf{NBA})~\cite{monti2021dag}, with the same splits and normalization procedure as in LED~\cite{mao2023leapfrog}. The dataset records 30 frames (6 seconds) of $(x,y)$ positions for 10 players and the ball. The prediction task is to observe the first 2 seconds (10 frames) and forecast the subsequent 4 seconds (20 frames).

\subsection{Implementation details}
In our set up, we use $S=50$ diffusion steps and $100$ epochs during training, with $\lambda$ values set to $0.01$. The batch size is $64$ for Basketball-U dataset and $16$ for the other three datasets. The learning rate is initialized to $10^{-3}$ and reduced by a factor of $2$ every $20$ epochs. The diffusion noise scheduler starts with an initial value of $\beta_0=10^{-4}$ and ends with $\beta_S=0.5$, following a quadratic function. The sampling is performed using hybrid DDIM, with a fixed skip interval of $\bar{\zeta}=10$ denoising steps, reducing the reverse process to only six denoising steps: $s \in \mathcal{D} = [50, 40, 30, 20, 10, 1]$. For U2Diff, the variance propagation starts at $\hat{s}=30$, whereas for U2Diffine, $\hat{s}=S$. The Temporal Mamba’s forward and reverse blocks are configured with a hidden size of $256$, matching the configuration of the Social Transformer, which uses a $1024$-dimensional feedforward layer and 8 attention heads. RankNN training involves generating $K=20$ modes per scene online using the trained U2Diff with frozen weights. The model is trained for $20$ epochs with a batch size of $32$ and a learning rate of $10^{-3}$, \changess{using the generated samples to compute rankings based on their proximity to ground truth. We also report a RankNN variant for U2Diffine generations; since online U2Diffine sampling is ${\sim}4\times$ more expensive, we fine-tune the U2Diff-trained checkpoint on U2Diffine samples for 5 epochs rather than retraining from scratch, which would be prohibitively slow. The architecture and loss are unchanged.}


\subsection{Evaluation metrics}
The first set of metrics are the commonly used the agent-wise metrics: minADE$_{K}$ as the minimum average displacement error, and minFDE$_{K}$ as the minimum final displacement error, both calculated over $K$ generated agent-modes. However, these metrics focus only on individual agent modes, lacking a full assessment of inter-agent scene dynamics. To address that, we include scene-level metrics: minSADE$_{K}$ as the minimum SADE (see~\eqref{eq:sade}), and minSFDE$_{K}$ as minimum scene final displacement error, both calculated over $K$ generated scene-modes~\cite{ngiam2021scene, girgis2021latent}.

We also adopt the metric used by~\cite{xu2025sportstraj} in trajectory completion evaluation, here referred to as minADE$_{K}$\cite{xu2025sportstraj}. The Spearman correlation coefficient $\rho$ (see~\eqref{eq:spearman}) is used to assess AvgUcty and RankNN operations. Finally, the Accuracy Rate (AccRate) metric evaluates uncertainty quality by measuring the percentage of ground-truth states that fall within the predicted distribution with 95\% confidence. This metric is computed using the Mahalanobis distance.

\begin{figure*}[t!]
\centering
\scalebox{0.94}{
\begin{tabular}{@{}ccccc@{}}
 UniTraj & Ours & UniTraj & Ours & Ours-20 \\ 
 
  \includegraphics[clip, angle=0,width=0.25\linewidth]{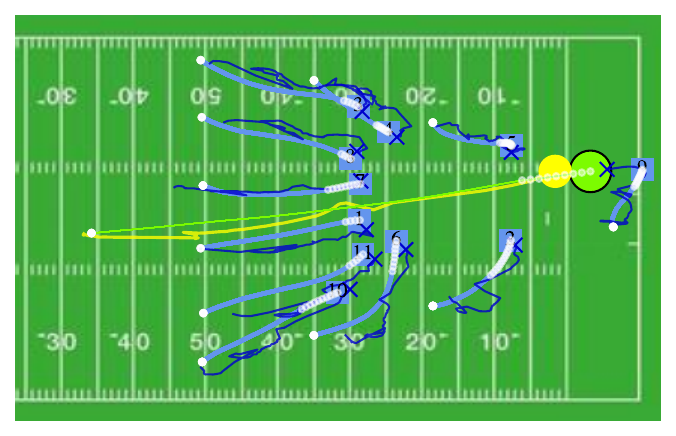}&
\hspace{-0.35cm}
\includegraphics[clip, angle=0,width=0.25\linewidth]{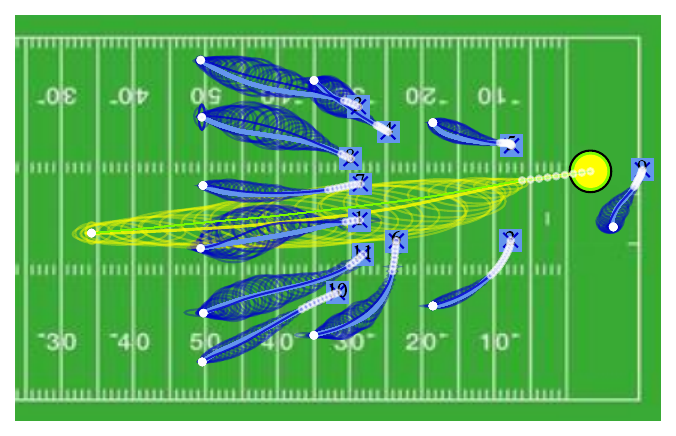}&
\hspace{-0.35cm}
  \includegraphics[clip, angle=0,width=0.165\linewidth, trim={0 0 7cm 0}]{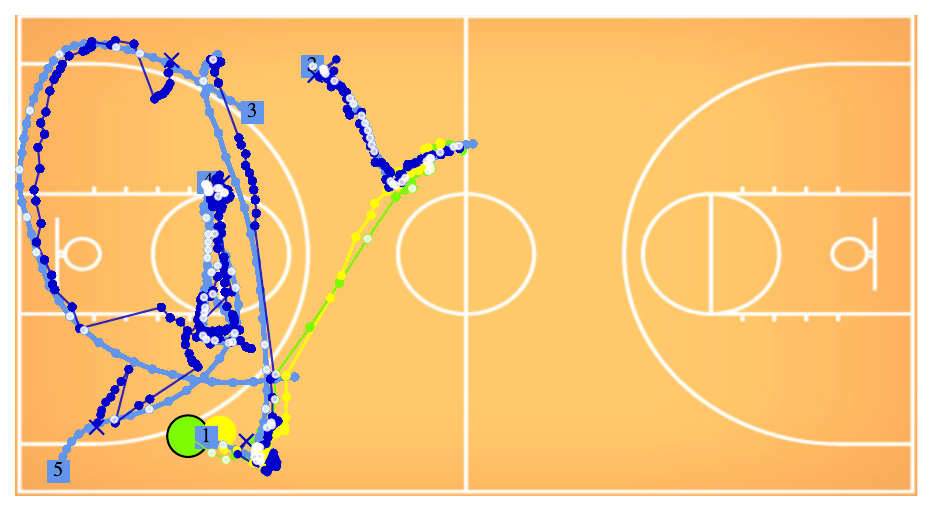}&
\hspace{-0.4cm}
\includegraphics[clip, angle=0,width=0.165\linewidth, trim={0 0 7cm 0}]{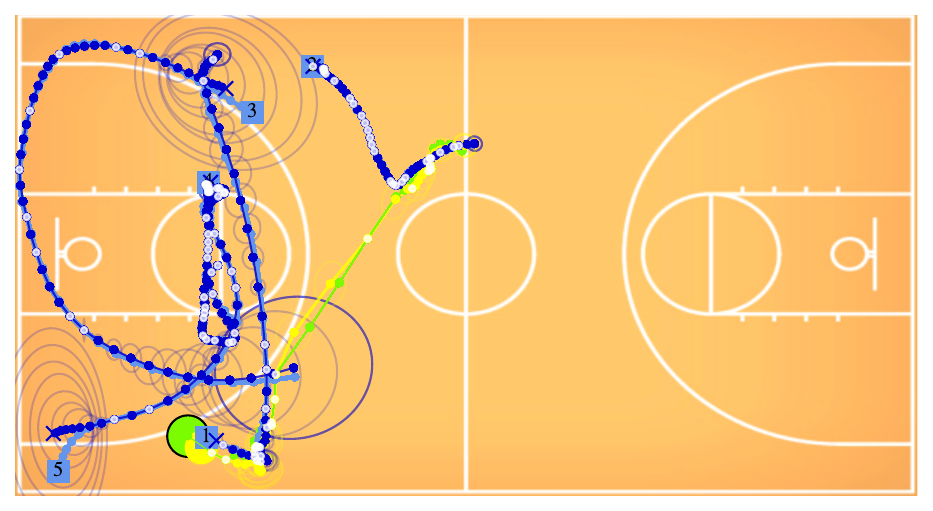}&
\hspace{-0.4cm}
\includegraphics[clip, angle=0,width=0.165\linewidth, trim={0 0 7cm 0}]{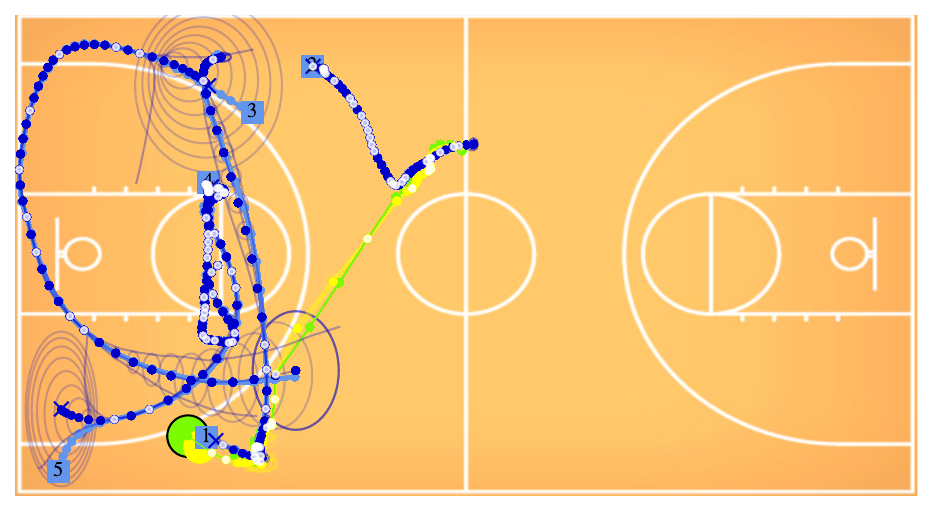}
\vspace{-0.2cm}\\

  \includegraphics[clip, angle=0,width=0.25\linewidth]{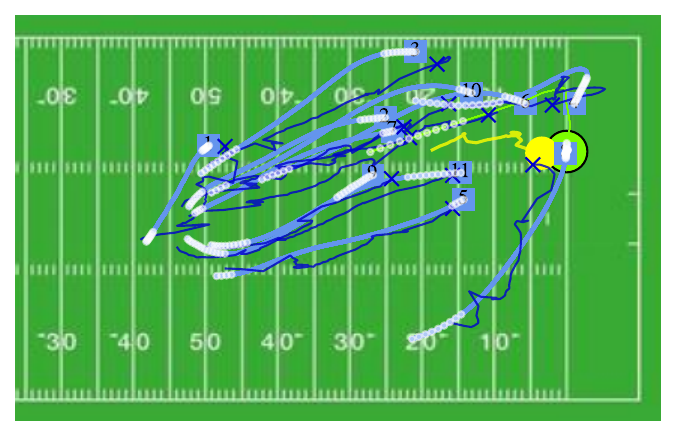}&
\hspace{-0.35cm}
\includegraphics[clip, angle=0,width=0.25\linewidth]{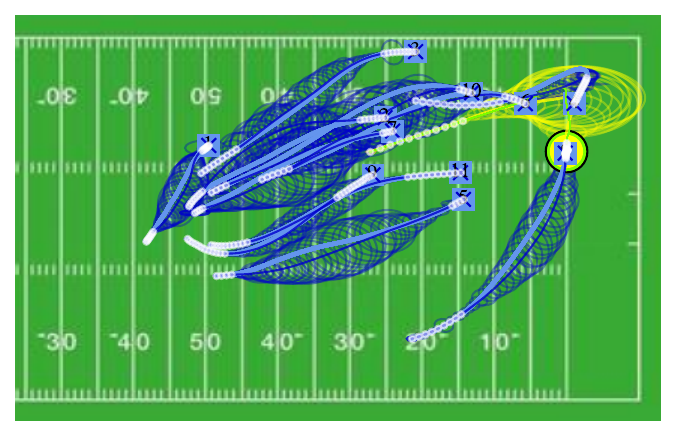}&
\hspace{-0.35cm}
  \includegraphics[clip, angle=0,width=0.165\linewidth, trim={0 0 7cm 0}]{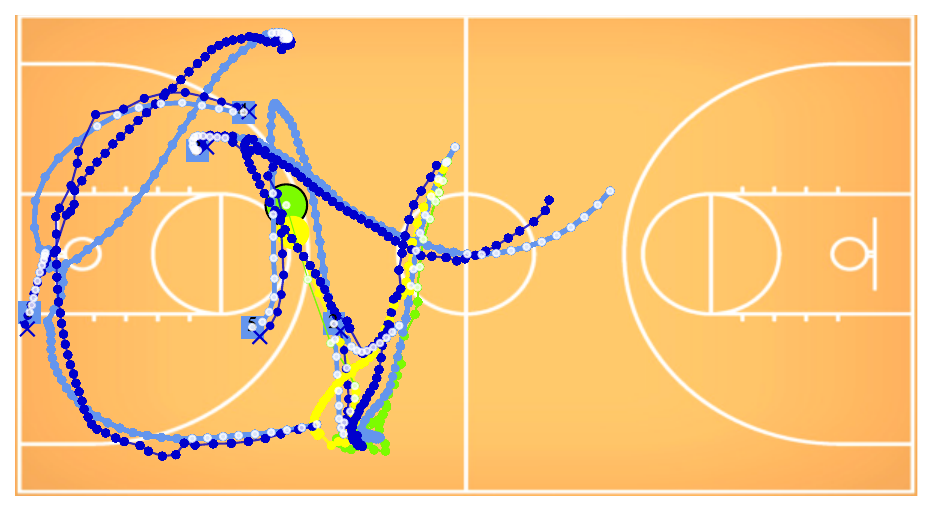}&
\hspace{-0.4cm}
  \includegraphics[clip, angle=0,width=0.165\linewidth, trim={0 0 7cm 0}]{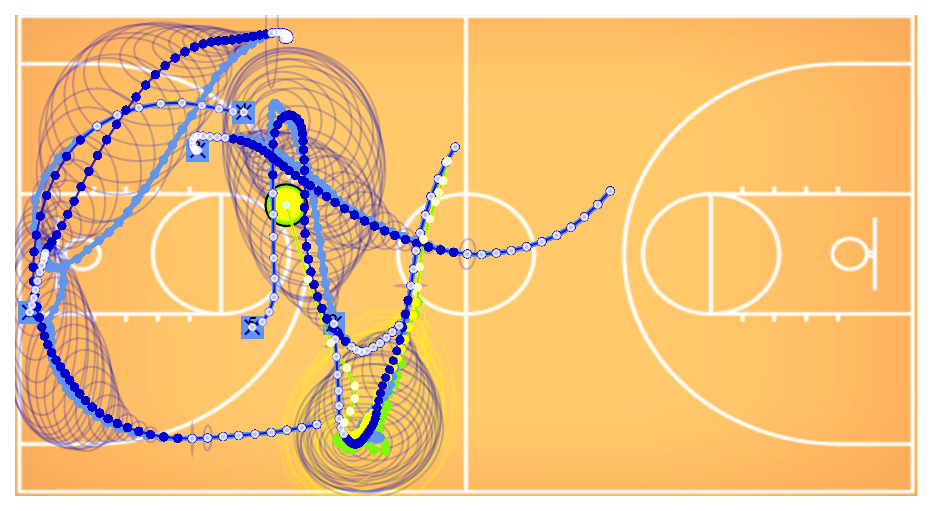}&
\hspace{-0.47cm}
\includegraphics[clip, angle=0,width=0.165\linewidth, trim={0 0 7cm 0}]{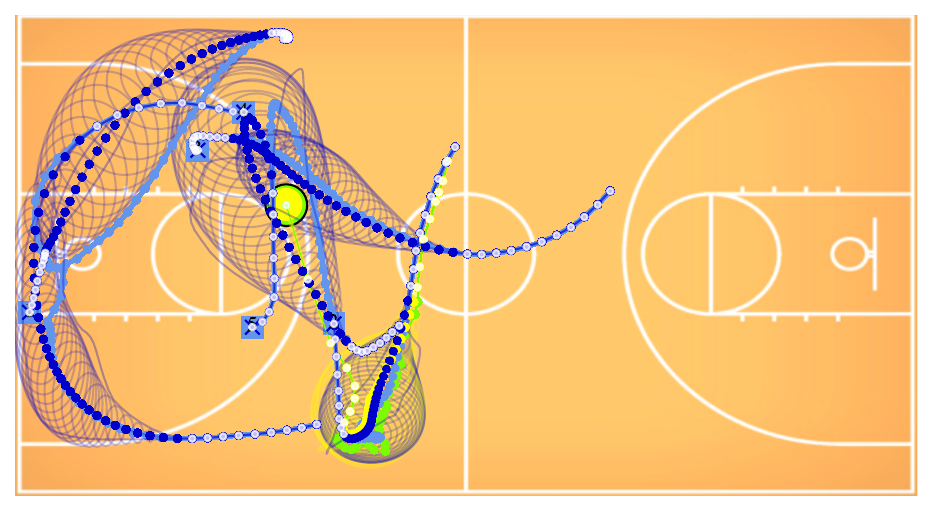}
\end{tabular}
}
\vspace{-0.2cm}
\label{fig:input_data}
\vspace{-0.2cm}
\end{figure*}
\begin{figure*}[t!]
\centering
\scalebox{0.94}{
\begin{tabular}{@{}cccc@{}}
 AutoBots & LED & MoFlow & Ours \\
 \includegraphics[clip, angle=0, width=0.25\linewidth]{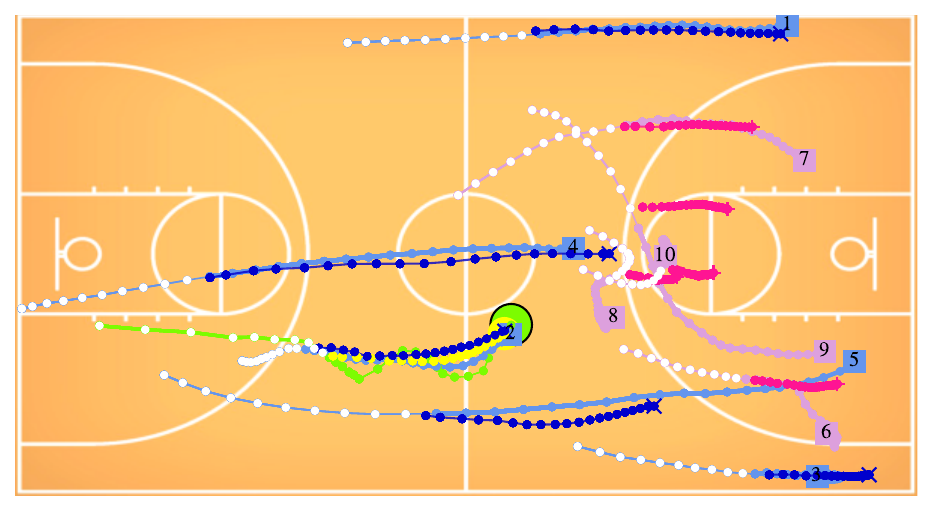}&
\hspace{-0.35cm}
 \includegraphics[clip, angle=0, width=0.25\linewidth]{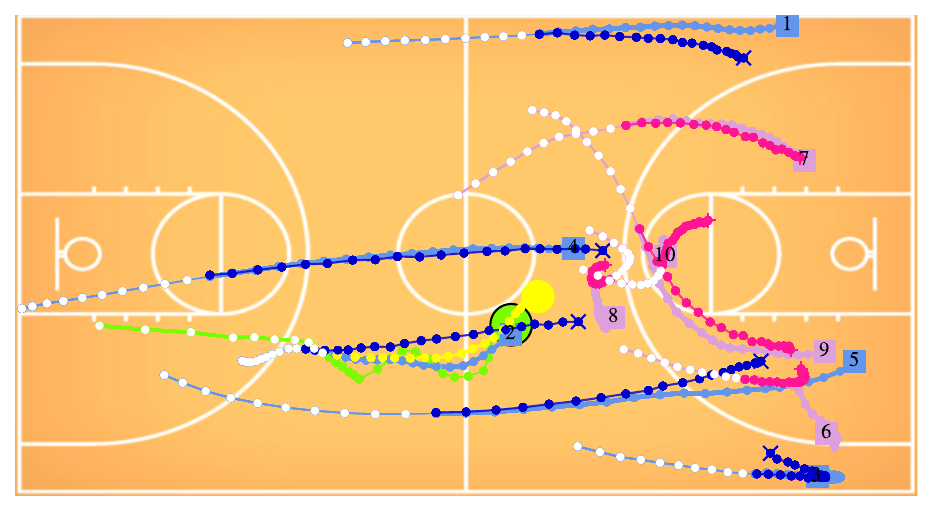}&
\hspace{-0.35cm}
 \includegraphics[clip, angle=0, width=0.25\linewidth]{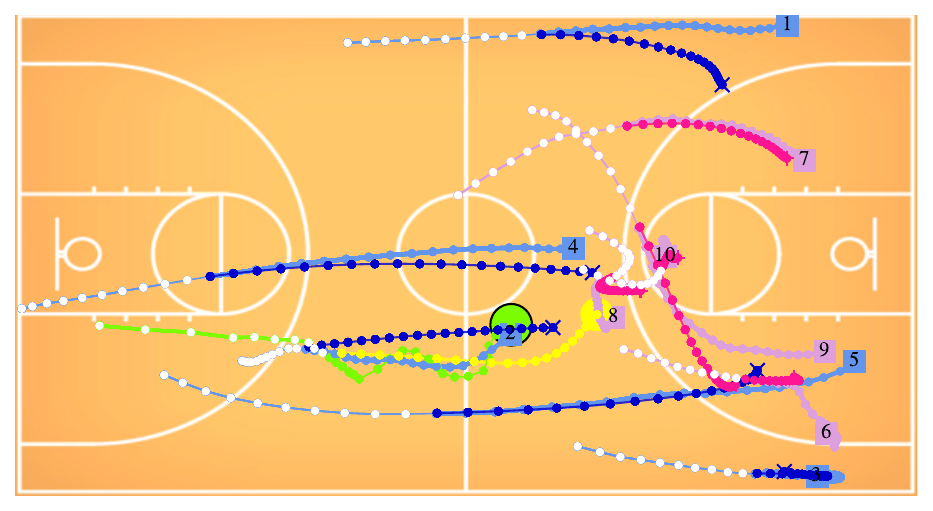}&
\hspace{-0.35cm}
 \includegraphics[clip, angle=0, width=0.25\linewidth]{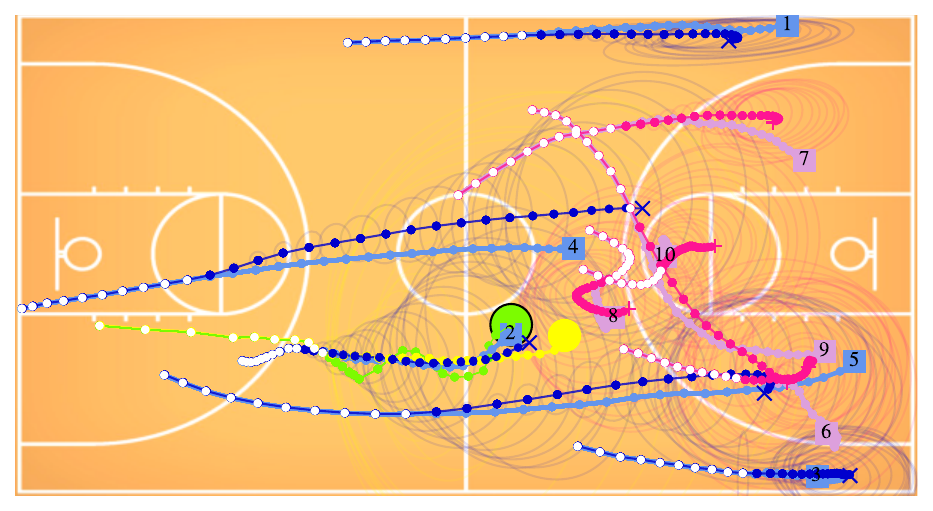}
\vspace{-0.2cm}\\
 \includegraphics[clip, angle=0, width=0.25\linewidth]{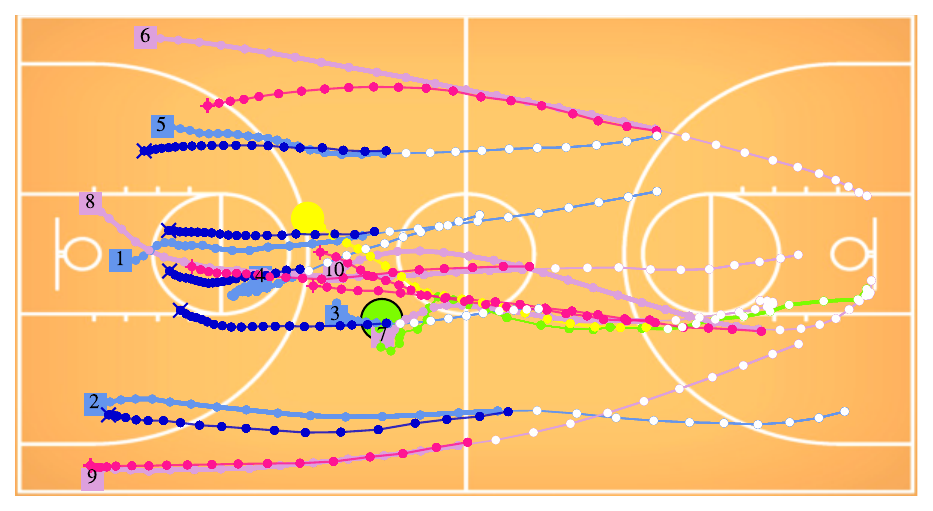}&
\hspace{-0.35cm}
 \includegraphics[clip, angle=0, width=0.25\linewidth]{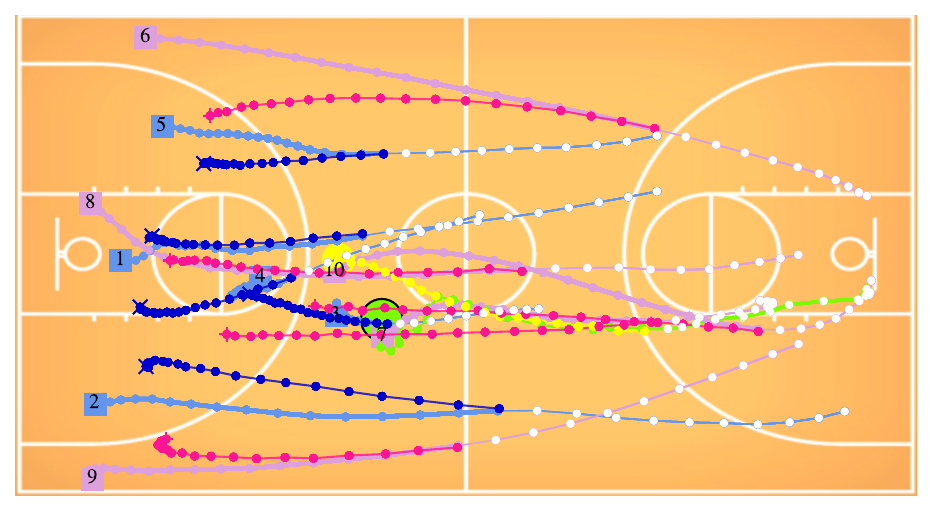}&
\hspace{-0.35cm}
 \includegraphics[clip, angle=0, width=0.25\linewidth]{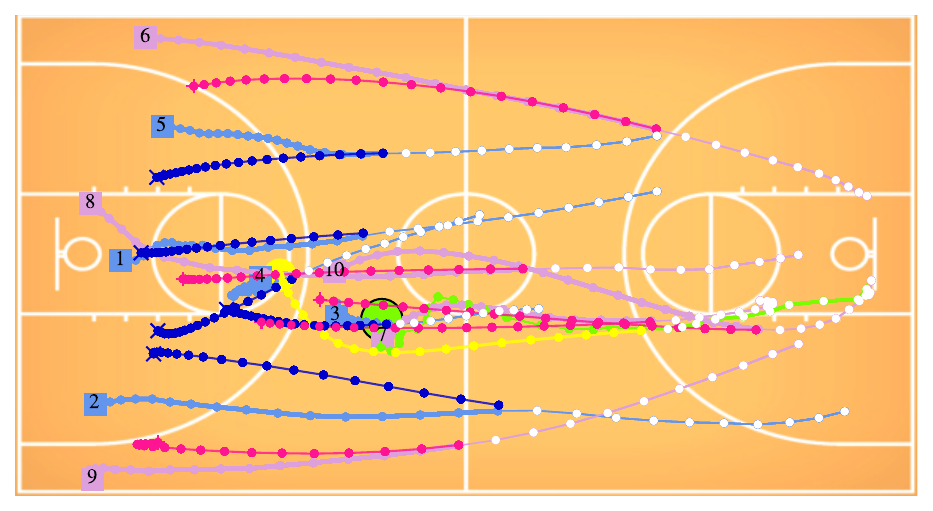}&
\hspace{-0.47cm}
 \includegraphics[clip, angle=0, width=0.25\linewidth]{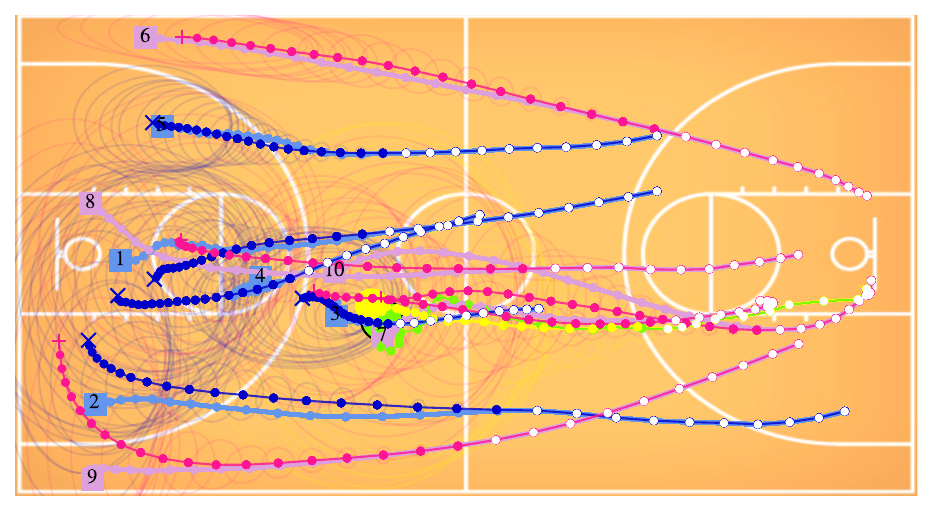}
\end{tabular}
}
\vspace{-0.2cm}
\caption{\textbf{Qualitative comparisons in trajectory completion (top) and forecasting (bottom).} \textbf{Top:} Comparison with UniTraj~\cite{xu2025sportstraj} on Football-U and Basketball-U datasets, and Ours-20 is the corresponding distribution over 20 generated modes. \textbf{Bottom:} Comparison against AutoBots~\cite{girgis2021latent}, LED~\cite{mao2023leapfrog} and MoFlow~\cite{fu2025moflow} on NBA trajectory forecasting. Ground truth player locations are shown in bright blue and pink, and the ball in green. Model input observations are in white. The predicted mode with the best minSADE$_{20}$ is shown, with players in dark blue and pink, and the ball in yellow.}
\label{fig:qualitative_results}
\vspace{-0.2cm}
\end{figure*}

\section{Experiments in trajectory modeling}
In this section, we compare our approach with several state-of-the-art methods in trajectory completion and trajectory forecasting tasks. It is important to note that RankNN is not evaluated in this section.

\begin{table}
\centering
\caption{\textbf{Evaluation on trajectory completion.} across three datasets. We report minADE$_{20}$~\cite{xu2025sportstraj}$\downarrow$ as the primary metric. For the UniTraj~\cite{xu2025sportstraj} baseline and our methods, we also include minSADE$_{20}$$\downarrow$.}
\scalebox{0.85}{
\begin{tabular}{lccc}
    \toprule
    Method & \multicolumn{1}{c}{Basketball-U (Feet)} & \multicolumn{1}{c}{Football-U (Yards)} & \multicolumn{1}{c}{Soccer-U (Pixels)} \\
    \midrule
    Mean & 14.58 & 14.18 & 417.68 \\
    Median & 14.56 & 14.23 & 418.06 \\
    Linear Fit & 13.54 & 12.66 & 398.34 \\   
    LSTM~\cite{hochreiter1997long} & 7.10 & 7.20 & 186.93 \\
    Transformer~\cite{vaswani2017attention} & 6.71 & 6.84 & 170.94 \\
    MAT~\cite{zhan2018generating} & 6.68 & 6.36 & 170.46 \\
    Naomi~\cite{liu2019naomi} & 6.52 & 6.77 & 145.20 \\ 
    INAM~\cite{qi2020imitative} & 6.53 & 5.80 & 134.86 \\
    SSSD~\cite{alcaraz2022diffusion} & 6.18 & 5.08 & 118.71 \\
    GC-VRNN~\cite{xu2023uncovering} & 5.81 & 4.95 & 105.87 \\
    UniTraj~\cite{xu2025sportstraj} & 4.77 (4.29) & 3.55 (4.03) & 94.59 (100.48) \\
    \midrule
    \textbf{U2Diff} & \underline{4.65} (\textbf{3.13}) & \underline{2.42} (\textbf{2.35}) & \underline{53.93} (\underline{51.14}) \\
    \textbf{U2Diffine} & \textbf{4.59} (\textbf{3.09}) & \textbf{2.38} (\textbf{2.35}) & \textbf{52.89} (\textbf{50.65}) \\

    \bottomrule
\end{tabular}
}
\label{tab:imputation}
\end{table}
\begin{table}
\centering
\caption{\textbf{Evaluation in trajectory forecasting} on NBA dataset. We report both agent-wise and scene-level metrics.}
\scalebox{0.89}{
\begin{tabular}{lccc}
    \toprule
    \multirow{2}{*}{Method} & \multicolumn{2}{c}{NBA (Meters)} \\
    \cmidrule(lr){2-3}
    & minADE$_{20}$ / minFDE$_{20}$ $\downarrow$ & minSADE$_{20}$ / minSFDE$_{20}$ $\downarrow$ \\
    \midrule
    MemoNet~\cite{xu2022remember}  & 1.15 / 1.57  & - \\
    NPSN~\cite{bae2022non}     & 1.25 / 1.47  & - \\
    GroupNet~\cite{xu2022groupnet}$^{\chi}$ & 0.94 / 1.22  & 2.12 / 3.72 \\
    AutoBots~\cite{girgis2021latent}$^{\psi}$ &  1.19 / 1.55 & 1.75 / 2.73  \\
    MID~\cite{gu2022stochastic}   & 0.96 / 1.27  &   - \\
    LED~\cite{mao2023leapfrog}   & 0.81 / 1.10  & 1.63 / 2.99 \\
    MART~\cite{lee2024mart}  &  \underline{0.72} / \underline{0.90}  & 1.52 / 2.77 \\
    MoFlow~\cite{fu2025moflow} & \textbf{0.71} / \textbf{0.86} & 1.52 / 2.73 \\
    \midrule
    \textbf{U2Diff} & 0.85 / 1.11 & \underline{1.48} / \underline{2.68}\\
    \textbf{U2Diffine} & 0.84 / 1.11 & \textbf{1.47} / \textbf{2.65}\\
    \bottomrule
    \multicolumn{4}{l}{\tiny $^{\chi}$ means a new pretrained model from their codebase is used, with better results than the reported in the original work.}\\
    \multicolumn{4}{l}{\tiny $^{\psi}$ means trained using their original code.} \\
\end{tabular}
}
\label{tab:forecasting}
\end{table}

In Table~\ref{tab:imputation}, we present the results for the minADE$_{20}$\cite{xu2025sportstraj} and, for UniTraj~\cite{xu2025sportstraj} and our baselines, the minSADE$_{20}$ metric (shown in parentheses). Our method outperforms UniTraj~\cite{xu2025sportstraj}, the strongest competing method, across all three completion datasets, achieving around 33\% and 44\% improvements in minADE$_{20}$~\cite{xu2025sportstraj} on the Football-U and Soccer-U datasets, respectively. When the number of agents is reduced, as in Basketball-U, we also obtain superior results, with a 27\% improvement in minSADE$_{20}$. The table also allows us to analyze the similar performance between U2Diff and U2Diffine in trajectory completion. It is shown that using bi-variate modeling (U2Diffine) in the loss function slightly improves the performance over uni-variate modeling (U2Diff). 

For the trajectory forecasting task, Table~\ref{tab:forecasting} reports results on the NBA dataset. Our unified approach ranks fourth on agent-level metrics (minADE$_{20}$/minFDE$_{20}$), while achieving state-of-the-art performance on scene-level metrics (minSADE$_{20}$/minSFDE$_{20}$), outperforming recent leading methods such as LED~\cite{mao2023leapfrog}, MART~\cite{lee2024mart}, and MoFlow~\cite{fu2025moflow} by more than 3\%. These baselines employ non-\textit{i.i.d.} mode-covering strategies, which are better suited for generating diverse samples and capturing out-of-distribution modes. In contrast, our model relies on a straightforward \textit{i.i.d.} sampling strategy, yet still establishes a new state-of-the-art. The sequential latent variable model AutoBots~\cite{girgis2021latent} also delivers competitive minSFDE$_{20}$ results. \changess{This table also includes U2Diff results; as in trajectory completion, U2Diffine's gain in displacement metrics is marginal, with its main benefit in uncertainty calibration, as it will be seen later.}

\vspace{-2mm}
Figure~\ref{fig:qualitative_results} illustrates examples for trajectory completion (top) and forecasting (bottom). The depicted trajectories are the modes with the minSADE$_{20}$. We compare in trajectory completion against the UniTraj~\cite{xu2025sportstraj}, where our method delivers more accurate reconstructions and plausible predictions in both Football-U and Basketball-U datasets. For NBA forecasting, our model provides superior future predictions compared to the state-of-the-art, especially in its ability to accurately model ball trajectories and their correlation with the ball carrier. This highlights the effectiveness of our scene-level metrics. Also note that our model is able to estimate variance in both tasks and reconstruct the observed states.

\subsection{Architecture ablation (U2Diffine)}

Table~\ref{tab:ablation_u2diff} reports ablation results on the architecture using the minSADE$_{20}$ metric across all datasets for the following configurations: ``w/o TM'', which replaces the Temporal Mamba with a Transformer Encoder to encode temporal dynamics, reverting to the original design in CSDI~\cite{tashiro2021csdi}; ``w/o ST'', which removes the Social Transformer, responsible for encoding agent interactions; ``w/o NLL'', considering $\lambda=0$ and consequently not minimizing the $\mathcal{L}_\text{NLL}$ loss function, resulting in $\mathcal{L}_\text{total} = \mathcal{L}_\text{simple}$.

The results highlight the importance of using Temporal Mamba instead of Transformers to significantly enhance performance compared to the original CSDI~\cite{tashiro2021csdi}. Furthermore, we demonstrate the crucial importance of encoding agents interactions with the Social Transformer in our multi-agent domain. The final loss term provides a slight improvement in performance while also giving our model the crucial ability to encode effective heteroscedastic uncertainty.

\begin{table}
\centering
\caption{\textbf{Ablation study} on U2Diff/U2Diffine architecture using the minSADE$_{20}\downarrow$ metric.}
\scalebox{0.94}{
\begin{tabular}{lcccc}
    \toprule
     \multirow{2}{*}{U2Diff}   &  Basketball-U & Football-U & Soccer-U & NBA  \\
     &  (Feet) & (Yards) & (Pixels) & (Meters)  \\
    \midrule
    w/o TM (CSDI~\cite{tashiro2021csdi})   &  3.74  & 2.70  & 58.78  &  1.73   \\
    w/o ST       &  4.06 & 4.75 & 96.66  &  1.92   \\
    w/o NLL & \underline{3.10}  & 2.37  & 51.27 & 1.50  \\
    \midrule
    \textbf{U2Diff}        & 3.13  & \textbf{2.35}  & \underline{51.14}  &  \underline{1.48}  \\
    \textbf{U2Diffine}  & \textbf{3.09} & \textbf{2.35} & \textbf{50.65} & \textbf{1.47}\\
    \bottomrule
\end{tabular}
}
\vspace{-1mm}
\label{tab:ablation_u2diff}
\vspace{-4mm}
\end{table}

\begin{table*}
\centering
\caption{\textbf{Ablation study on reverse Gaussian sampling.} We report NLL$\downarrow$,  AccRate$\uparrow$ (\%), and the Spearman Correlation $\rho \uparrow$. Values in parentheses indicate the minimum over $K=20$ modes. \textcolor{red}{U2Diff} and \textcolor{blue}{U2Diffine} are highlighted. \vspace{-1mm}} 
\scalebox{0.90}{
\begin{tabular}{lcccccccccccc}
    \toprule
    \multirow{2}{*}{\makecell{Reverse \\ Gaussian}} &  \multicolumn{3}{c}{Basketball-U} & \multicolumn{3}{c}{Football-U} & \multicolumn{3}{c}{Soccer-U} & \multicolumn{3}{c}{NBA}  \\
    \cmidrule(lr){2-4} \cmidrule(lr){5-7} \cmidrule(lr){8-10} \cmidrule(lr){11-13}
      & NLL$\downarrow$ & AccRate$\uparrow$ & $\rho\uparrow$ & NLL$\downarrow$ & AccRate$\uparrow$ & $\rho\uparrow$ & NLL$\downarrow$ & AccRate$\uparrow$ & $\rho\uparrow$ & NLL$\downarrow$ & AccRate$\uparrow$ & $\rho\uparrow$ \\
    \midrule
    \textcolor{red}{uni / $\bmJ_s = 0$}              & -1.24 (-1.76) & 85.9 (90.4) &  0.27 &  -2.16 (-2.39) & 93.8 (95.5) & 0.28 & -2.51 (-2.66) & 95.3 (96.8) & 0.23 & -0.94 (-1.31) & \textbf{84.3} (\textbf{91.6})  & \textbf{0.30}\\
    uni / $\bmJ_s$   & \textbf{-2.01} (\underline{-2.59})  & 86.3 (91.0) &  \textbf{0.31}  & \textbf{-3.22} (\textbf{-3.32}) & 95.0 (96.6) & 0.33 & \textbf{-3.60} (\underline{-3.70}) & \textbf{97.3} (\textbf{98.4}) & 0.35 & \textbf{-1.10} (\textbf{-1.70}) & 80.9 (87.8) & 0.26 \\
    bi / $\bmJ_s = 0$       & -1.22 (-1.79)  & 86.6 (91.0) & 0.25 & -2.12 (-2.34) & 93.8 (95.5) & 0.30 & -2.55 (-2.71) & 95.3 (96.8) & 0.34 & -0.95 (-1.34) & \underline{83.3} (\underline{91.0}) & \underline{0.29} \\
    bi / $\bmJ_s$ full             & -1.97 (\underline{-2.59}) & \textbf{88.0} (\textbf{91.9}) &  0.28 &  -3.18 (-3.29) & \textbf{95.1} (\textbf{96.7}) & \underline{0.35} & -3.58 (-3.69) & \underline{97.0} (\underline{98.3}) & \underline{0.37} & -1.09 (-1.69) & 81.3 (87.8)  & 0.27 \\
    \textcolor{blue}{bi / $\bmJ_s$}        & \underline{-2.00} (\textbf{-2.62})  & \underline{87.9} (\textbf{91.9})  & \underline{0.30} & \textbf{-3.22} (\textbf{-3.32}) & \textbf{95.1} (\textbf{96.7}) & \textbf{0.37} & \textbf{-3.60} (\textbf{-3.71}) & \underline{97.0} (\underline{98.3}) & \textbf{0.38} & \textbf{-1.10} (\textbf{-1.70}) & 81.1 (87.7) & \underline{0.29}  \\
    \cmidrule(lr){1-1} \cmidrule(lr){2-4} \cmidrule(lr){5-7} \cmidrule(lr){8-10} \cmidrule(lr){11-13}
     & \multicolumn{3}{c}{Inference time$\downarrow$ (ms)} & \multicolumn{3}{c}{Inference time$\downarrow$ (ms)} & \multicolumn{3}{c}{Inference time$\downarrow$ (ms)} & \multicolumn{3}{c}{Inference time$\downarrow$ (ms)}  \\
     $\cdot$ / $\bmJ_s=0$& \multicolumn{3}{c}{08 (x1.00)} & \multicolumn{3}{c}{14 (x1.00)} & \multicolumn{3}{c}{14 (x1.00)} & \multicolumn{3}{c}{06 (x1.00)}  \\
     $\cdot$ / $\bmJ_s$& \multicolumn{3}{c}{31 (x3.88)} & \multicolumn{3}{c}{59 (x4.21)} & \multicolumn{3}{c}{59 (x4.21)} & \multicolumn{3}{c}{23 (x3.83)}  \\
    
    \bottomrule
\end{tabular}
}
\label{tab:ablation_u2diff_nll}
\vspace{-4mm}
\end{table*}

\section{Uncertainty and error probability estimation}
We now evaluate our approach's ability to compute accurate variances for each state, as well as the error probability derived from RankNN.

\begin{figure*}
  \centering
  \includegraphics[width=1.0\linewidth]{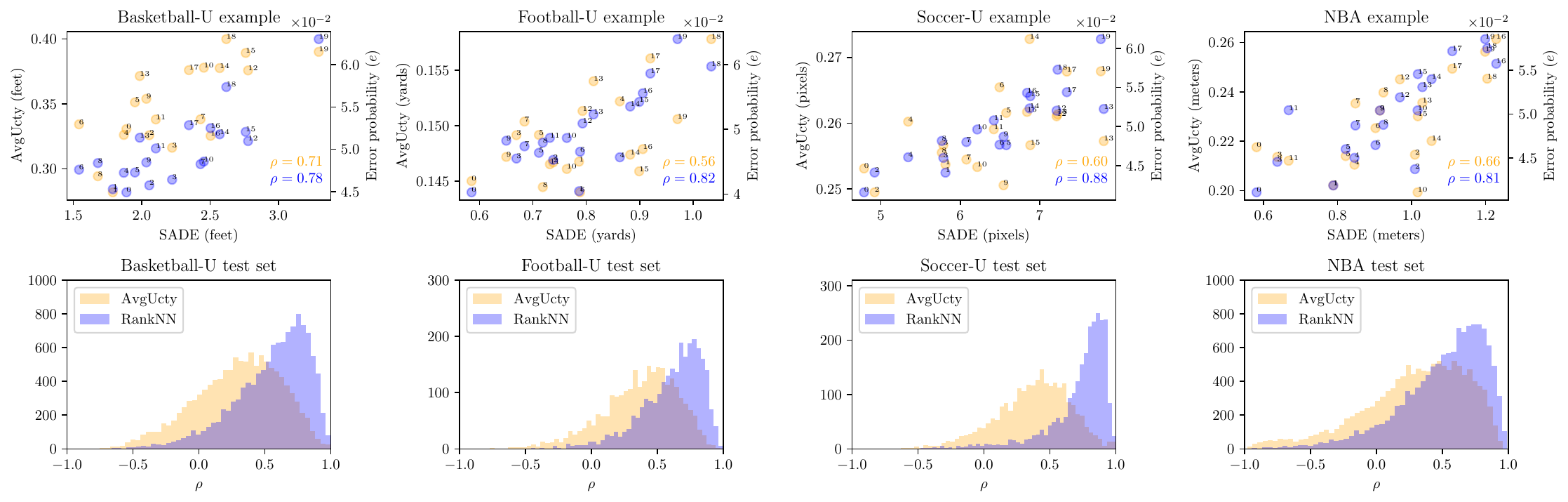}
  \vspace{-8mm}
\caption{\textbf{Qualitative evaluation of the error correlation.} \textbf{Top:} In orange, the AvgUcty versus SADE across the 20 generated modes using U2Diffine of a test scene example. In blue, the error probability $e$ versus SADE. \textbf{Bottom:} Distribution of Spearman correlation coefficients $\rho$ for all four test datasets, using AvgUcty in orange and RankNN predicting $e$ in blue.}
  \label{fig:scatter_histograms}
  \vspace{-3mm}
\end{figure*}

Figure~\ref{fig:qualitative_results} illustrates the predicted uncertainty for each state, visualized as a 95\% confidence interval. For trajectory completion, our model demonstrates robust uncertainty estimation: the ground-truth states consistently fall within the predicted \changess{uncertainty ellipse} in almost all instances. However, NBA forecasting remains more challenging due to the difficulty in maintaining high confidence over longer prediction horizons. What is notable with U2Diffine, though, is its bi-variate uncertainty modeling. This feature allows the model to predict rotated uncertainty ellipses, which closely align with the spatial distribution of the data. Additionally, we have included the Ours-20 variant. Following CSDI~\cite{tashiro2021csdi} idea to represent the Gaussian process, Ours-20 represents the distribution of predictions derived from 20 sampled trajectories. \changes{The increased spread in regions far from observed states demonstrates that the model generates diverse, plausible modes rather than collapsing to a single prediction. This visualization additionally depicts upper and lower confidence bounds for each player, highlighting variability across modes. A supplementary video further illustrates the diversity of generated modes.} By comparing Ours (our model's predicted variance) with Ours-20 (the sampling approximation), we highlight how effectively our approach captures heteroscedastic uncertainty in trajectory predictions, showcasing a strong alignment between the two.

We quantitatively evaluate the uncertainty using three key metrics: Negative Log-Likelihood (NLL), Accuracy Rate (AccRate), and Spearman Correlation ($\rho$) between the AvgUcty and the SADE error. Table~\ref{tab:ablation_u2diff_nll} extends Fig.~\ref{fig:ablation_jacobian} (see section~\ref{subsec:rev_gaussian}) presenting a comparative analysis of bi-variate versus uni-variate modeling (``bi / $\cdot$ '' vs ``uni / $\cdot$'') , and the impact of computing or not computing gradients (`` $\cdot$ / $\bmJ_s$'' vs `` $\cdot$ / $\bmJ_s=\bm0$''). The metrics were computed as an average over $K=20$ generated modes. For NLL and AccRate, the minimum value across all modes is also included in parentheses. \changes{For completeness, we add the results using the full form of the Jacobian (``bi / $\bmJ_s$ full"). As previously mentioned, it produces numerical instabilities in some rare states, therefore we curate the computed Jacobian by clamping its singular values in a range $[-100, 100]$.} 


Our U2Diffine model demonstrates robust variance estimation, achieving over 95\% AccRate on Football-U and Soccer-U datasets. While Basketball-U and NBA forecasting prove more challenging with lower AccRate and higher NLL due to increased dynamics and longer horizons, our method consistently performs well. As shown in the table, computing gradients (`` $\cdot$ / $\bmJ_s$'') consistently boosts NLL performance. The $\rho$ metric remains consistent, with our U2Diffine baseline (``bi / $\bmJ_s$'') achieving moderate correlations (0.29--0.38). While generally a sub-optimal variant, our U2Diff baseline (``uni / $\bmJ_s=\bm0$'') achieves the best AccRate and $\rho$ results on the NBA dataset. To provide insight into computational overhead, we report the inference time (in milliseconds per mode) for a batch of 128 samples, each with 20 generated modes, highlighting the impact of gradient computation. \changes{The full variant (``bi / $\bmJ_s$ full'') shows slightly worse NLL and $\rho$, suggesting more conservative uncertainty estimates due to the Jacobian post-processing.} Computationally, U2Diffine's gradient calculation requires approximately 4 times the inference time of U2Diff on an RTX A6000 GPU.  However, this is still significantly more efficient than a Bayesian Laplace approximation method like the one proposed in~\cite{kou2023bayesdiff}, which would require 10 times the inference time of U2Diff even with a small sampling size. We note that we did not report results for this method as it did not converge within our multi-agent trajectory completion framework.

As stated previously, AvgUcty moderately correlates with real-scene closeness. However, this correlation significantly improves with the RankNN approach. Table~\ref{tab:spearmancorr} presents mean and median Spearman correlation ($\rho$) values for AvgUcty and predicted error probability ($e$) across all datasets and scenes, including an AutoBots~\cite{girgis2021latent} comparison for the NBA dataset. Our predicted state-wise uncertainty and estimated error probabilities consistently achieve higher correlations. Specifically, U2Diffine baseline's median AvgUcty $\rho$ values range from 0.32 to 0.41. For error probability ($e$), median results boost to 0.58 to 0.79. Results are similar between our two proposed baselines, possibly indicating a performance plateau or being attributable to U2Diffine's fine-tuning settings. Figure~\ref{fig:scatter_histograms}-top visually illustrates these findings with four U2Diffine examples, comparing AvgUcty and error probabilities ($e$) against SADE for each mode under the same prior. Blue dots, representing error probabilities, clearly show superior rankings over AvgUcty. The full distribution of Spearman correlations across test sets for U2Diffine is presented in Fig.~\ref{fig:scatter_histograms}-bottom.

\begin{table}
\centering
\caption{\textbf{Evaluation of the Spearman Correlation with error.} The results are the Mean / Median of $\rho \uparrow$ between the rank method and the SADE.}
\scalebox{0.80}{
\begin{tabular}{lccccc}
    \toprule
    Method &  Rank  & Basketball-U & Football-U & Soccer-U & NBA  \\
    \midrule
    \multirow{2}{*}{AutoBots~\cite{girgis2021latent}}  & AvgUcty & - & - & - & 0.09 / 0.10  \\
     &  $e$   &  - & - & - & 0.37 / 0.44  \\
    \midrule
    \multirow{2}{*}{\textbf{U2Diff}} & AvgUcty & 0.27 / 0.29 & 0.28 / 0.31 & 0.23 / 0.25 & 0.30 / 0.35 \\  
      & $e$   &  \textbf{0.56} / \textbf{0.63}   &  \underline{0.59} / \underline{0.65}  & \textbf{0.72} / \underline{0.78} &  \textbf{0.51} / \textbf{0.58}   \\
    \midrule
    \multirow{2}{*}{\textbf{U2Diffine}} & AvgUcty &  0.29 / 0.32 & 0.37 / 0.41 & 0.38 / 0.40 & 0.29 / 0.35 \\  
      & $e$   & \underline{0.55} / \underline{0.61} & \textbf{0.61} / \textbf{0.66} & \textbf{0.72} / \textbf{0.79} & \textbf{0.51} / \textbf{0.58}    \\
    \bottomrule
\end{tabular}} 
\label{tab:spearmancorr}
\vspace{-3mm}
\end{table}

Ranking strategies can also be evaluated by their impact on trajectory accuracy. Table~\ref{tab:comparison_topk} shows results for different ranking strategies used to select Top-$k$ modes from a set of 20, then computing minSADE$_{k}$. Note that Top-20 is equivalent to minSADE$_{20}$. The considered ranking methods include Random, AvgUcty (for models outputting variance), and error probability (e) ranking. We present results for LED~\cite{mao2023leapfrog}, MART~\cite{lee2024mart}, MoFlow~\cite{lee2024mart}, and AutoBots~\cite{girgis2021latent}. Our findings clearly show that using AvgUcty and RankNN probabilities, our methods consistently outperform the state-of-the-art, significantly improving forecasting accuracy. Notably, simple AvgUcty on predicted variance reduces Top-1 error from 2.01 to 1.91 (U2Diff) and to 1.93 (U2Diffine). Integrating the post-processing RankNN framework to infer $e$ further reduces this error to 1.82 (U2Diff) and 1.81 (U2Diffine).

\begin{table}
\centering
\caption{\textbf{Evaluation of the rank techniques and baselines comparisons.} From 20 generated modes, the Top-$k$ ranked by each method are selected, and minSADE$_k\downarrow$ is computed over this subset.}
\scalebox{0.85}{
\begin{tabular}{lcccccc}
    \toprule
    \multirow{2}{*}{Method} & \multirow{2}{*}{Rank} & \multicolumn{5}{c}{NBA (Meters)} \\
    \cmidrule(lr){3-7}
     & &  Top-1  & Top-3  & Top-5  &  Top-10  & Top-20\\
    \midrule
    LED~\cite{mao2023leapfrog} & Random              & 3.80  &  2.17 &  1.92  &  1.73  &  1.63    \\
    MART~\cite{lee2024mart} & Random & 2.46 & 1.88 & 1.73 & 1.60 & 1.52 \\
    MoFlow~\cite{fu2025moflow} & Random & 2.41 & 1.87 & 1.73 & 1.60 & 1.52 \\
    \midrule
    \multirow{3}{*}{AutoBots~\cite{girgis2021latent}} & Random        & 2.76  &  2.17 &  2.02  &  1.88  &  1.75\\
     & AvgUcty & 2.37  &  2.19  & 2.09  &  1.94  &  1.75   \\
     & $e$   & 2.40  &  2.08  & 1.98  &  1.86  &  1.75   \\
     \midrule     
     \multirow{3}{*}{\textbf{U2Diff}}    & Random          & 2.01  &  1.75  & 1.66 &  1.56  &  1.48     \\
       & AvgUcty &  1.91  &  1.71  & 1.63    & 1.55   & 1.48      \\
      & $e$   &  \underline{1.82}   &  \textbf{1.66}  &  \textbf{1.60}   & \underline{1.54}   &  \underline{1.48}     \\
      \midrule     
      \multirow{3}{*}{\textbf{U2Diffine}}    & Random          & 2.01  &  1.74  & 1.65 &  1.55  &  1.47     \\
       & AvgUcty &  1.93  &  1.72  & 1.64    & 1.54   & 1.47      \\
      & $e$   &  \textbf{1.81} & \textbf{1.66} & \textbf{1.60} & \textbf{1.53} & \textbf{1.47}    \\
    \bottomrule
\end{tabular}
}
\label{tab:comparison_topk}
\end{table}
\begin{table}
\centering
\caption{\textbf{Ablation study on RankNN using the Mean of Spearman correlation coefficient ($\rho\uparrow$)}. The results are presented across all evaluated datasets.}
\scalebox{0.95}{
\begin{tabular}{lcccc}
    \toprule
     RankNN   &  Basketball-U & Football-U & Soccer-U & NBA  \\
    \midrule
    w/o TM w/o ST   &  0.28 & 0.37 & 0.31 & 0.40 \\
    w/o TM          &  0.35 & 0.50 & 0.54 & 0.45 \\
    w/o ST          &  0.39 & 0.51 & 0.34 & 0.45 \\
    w/o MST         &  0.52 & 0.47 & 0.65 & 0.50 \\
    w/o VAR         &  0.54 & 0.56 & 0.55  &  0.50 \\  
    \midrule
    \textbf{U2Diff}            & \textbf{0.56}  &  \underline{0.59} & \textbf{0.72}  &  \textbf{0.51} \\
    \textbf{U2Diffine} & \underline{0.55} & \textbf{0.61} & \textbf{0.72} & \textbf{0.51} \\
    \bottomrule
\end{tabular}
}
\vspace{-1mm}
\label{tab:ablation_ranknn}
\vspace{-4mm}
\end{table}

\subsection{Architecture ablation (RankNN)}
We perform an ablation study on the RankNN architecture using U2Diff online generated modes. Results are presented in Table~\ref{tab:ablation_ranknn}, using the Spearman correlation coefficient ($\rho$) as the evaluation metric to measure the alignment between $e$ and SADE. The study evaluated the following configurations: ``w/o TM w/o ST'' skips the Social-temporal Block entirely, bypassing both Temporal Mamba and Social Transformer processing; ``w/o TM'' excludes the Temporal Mamba processing, while retaining the Social Transformer; ``w/o ST'' excludes the Social Transformer processing, while retaining the Temporal Mamba; ``w/o MST'' skips the Multi-scene Transformer processing; and ``w/o VAR'' removes the predicted variance $\Var(\bmXh_0)$ from the input to RankNN, using only the mean predicted locations $\bmX_0$ and the binary mask $\mathbf{M}$. The last configuration reduces the input tensor from $K \times T \times N \times 5$ to $K \times T \times N \times 3$.  

The results emphasize the importance of processing states through the Social-Temporal Block, demonstrating the critical contributions of both the Temporal Mamba and the Social Transformer. Furthermore, the w/o MST configuration leads to suboptimal performance, particularly on the Football-U dataset, underscoring the value of Multi-scene Transformer processing. Finally, the w/o VAR configuration highlights the role of the predicted variance $\Var(\bmXh_0)$.

\section{Conclusion}
This paper introduces a unified uncertainty-aware diffusion framework for general trajectory completion. Our method not only surpasses state-of-the-art forecasting baselines in scene-level metrics but also establishes a new benchmark for trajectory completion tasks. We demonstrate its effectiveness in estimating state-wise uncertainty through a novel loss augmentation. We propose two approximations for uncertainty computation: U2Diffine, which uses a first-order Taylor approximation via Reverse Gaussian Sampling and requires gradient computation; and U2Diff, a faster approximation that avoids gradient computation. Both baselines exhibit state-wise variance with a stronger correlation to ground truth errors than scene-level state-of-the-art methods. Furthermore, we present a novel post-processing supervised RankNN model that infers error probability estimates for each mode, achieving high correlation with ground truth errors and outperforming scene-level based approaches.

\noindent \textbf{Acknowledgment.} This work has been supported by the project GRAVATAR PID2023-151184OB-I00 funded by MCIU/AEI/10.13039/501100011033 and by ERDF, UE and by the Government of Catalonia under 2023 DI 00058.

{
    \small
    \bibliographystyle{IEEEtran}
    \bibliography{main}
}

\begin{IEEEbiography}[{\includegraphics[width=1in,height=1.25in,clip,keepaspectratio]{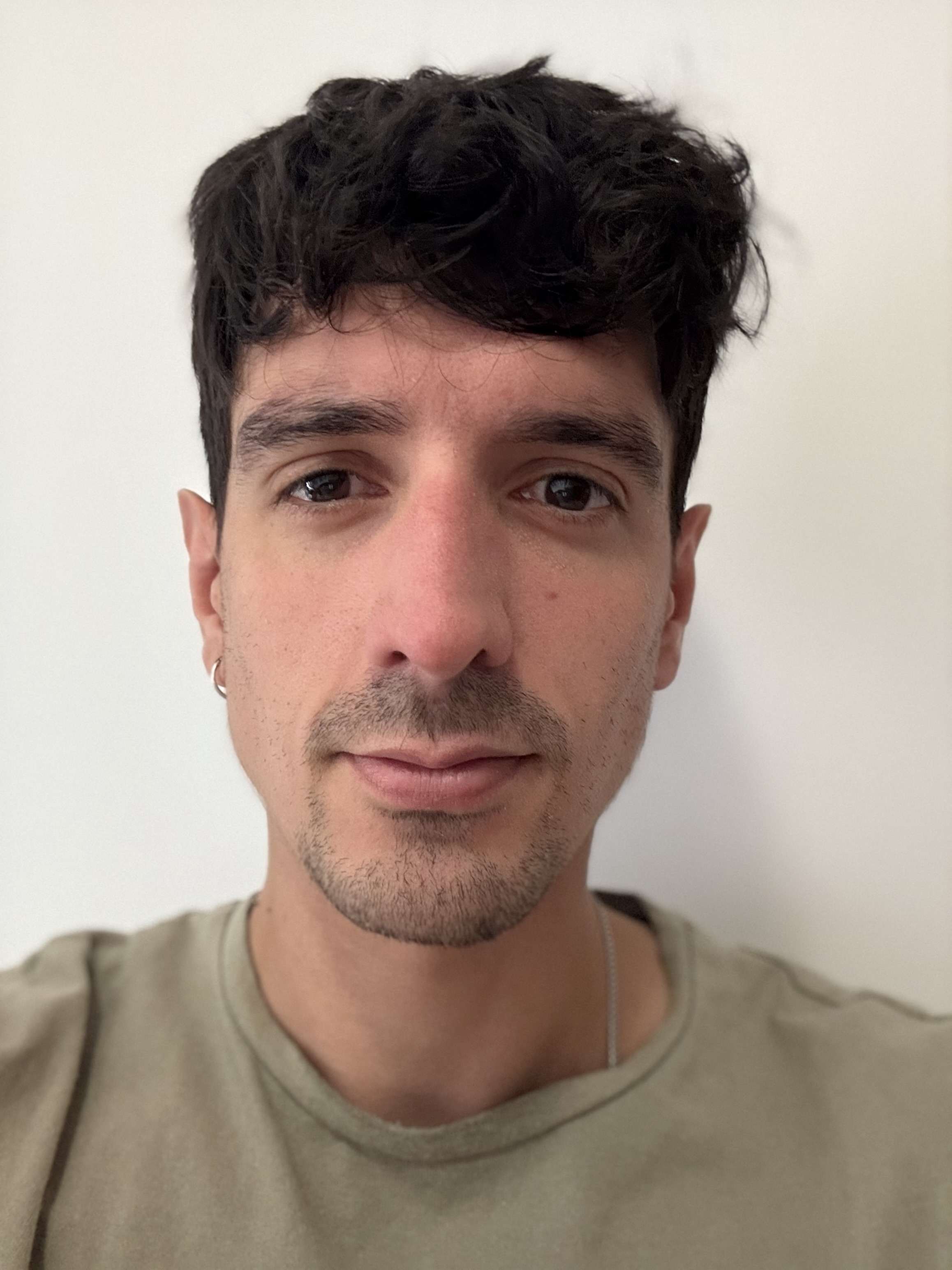}}]{Guillem Capellera}
received the B.Sc. degree in Sports Science in 2017, B.Sc. degree in Mathematics and Physics in 2021, from University of Barcelona (UB); he then earned an M.Sc. degree in Computer Vision from Autonomous University of Barcelona (UAB). He began an industrial Ph.D. in Automatic Control, Robotics and Vision in 2023, carried out jointly at the Institut de Robòtica i Informàtica Industrial (CSIC-UPC) and the company Kognia Sports Intelligence, with focus on modeling multi-agent trajectories in sports. He conducted a research stay at the Visual Intelligence for Transportation (VITA) group of Ecole Polytechnique Fédérale de Lausanne (EPFL) in 2025.
\end{IEEEbiography}
\vspace{-0.6cm}
\begin{IEEEbiography}[{\includegraphics[width=1in,height=1.25in,clip,keepaspectratio, trim={2cm 0 2cm 0}]{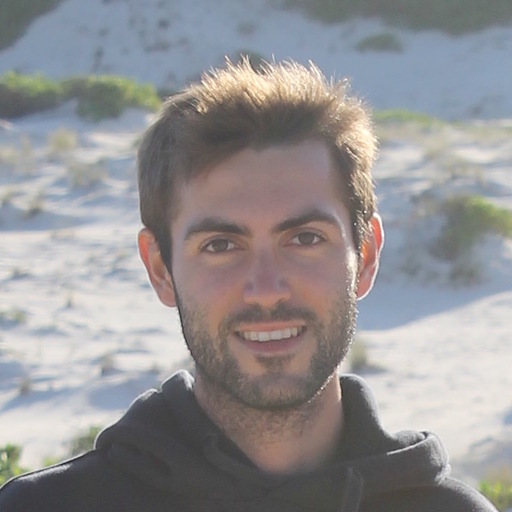}}]{Antonio Rubio}
obtained his M.Sc. degree in Automatic Control and Robotics from the Universitat Politècnica de Catalunya (UPC) in 2015. He completed an Industrial Ph.D. in Computer Vision in 2021, carried out jointly at the Institut de Robòtica i Informàtica Industrial (CSIC-UPC) and the company Wide Eyes, with a focus on multi-modal embedding techniques. In 2017, he conducted a research stay at Waseda University in Tokyo, and in 2019, he held a teaching position in the Biomedical Engineering program at the Universitat de Vic (UVic-UCC). He was awarded the Best Paper Award at the ICCV 2017 CVF Workshop. Since 2019, he has been working at Kognia Sports Intelligence, where he develops video-based algorithms for football match analysis, while also supervising Master’s and Ph.D. students.
\end{IEEEbiography}
\vspace{-0.6cm}
\begin{IEEEbiography}[{\includegraphics[width=1in,height=1.25in,clip,keepaspectratio]{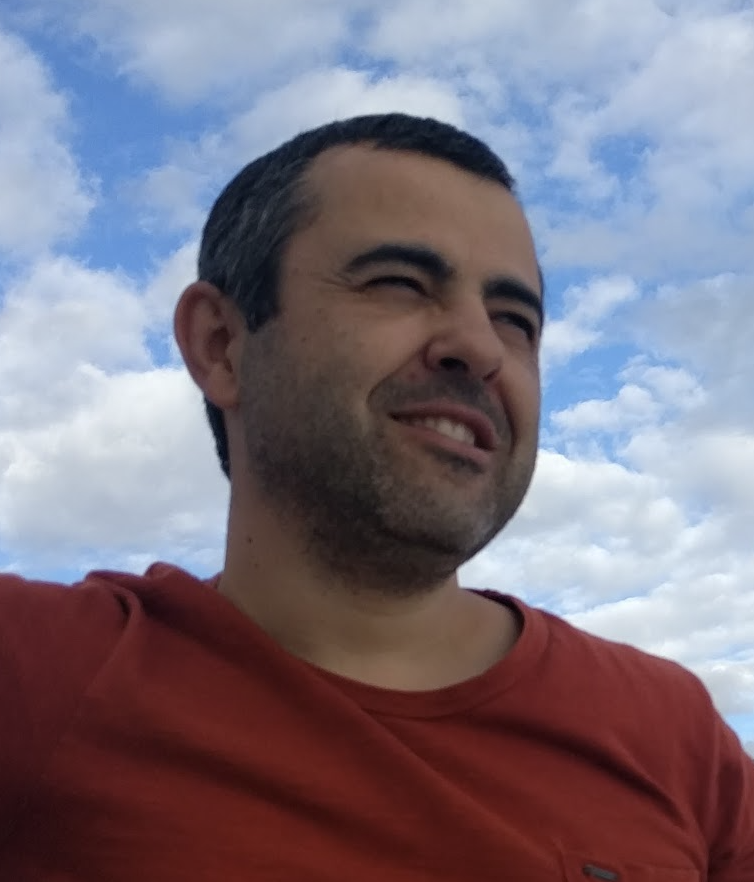}}]{Luis Ferraz}
received his M.Sc. in Computer Vision and Artificial Intelligence from the Universitat Autònoma de Barcelona in 2007. He completed an Executive Master's degree in Innovation from the Escuela de Organización Industrial in 2015. In 2016, he earned his Ph.D. in Computer Vision and Artificial Intelligence from the Universitat Autònoma de Barcelona focused on viewpoint invariant features and robust monocular camera pose estimation. Since 2019, he is the Head of AI at Kognia Sports intelligence, leading a multidisciplinary team dedicated to leveraging advanced AI to unravel the complexities of soccer. He is particularly interested in extracting high-level semantics from spatio-temporal data such as tactical patterns and the contextual meaning of player actions to provide human-interpretable insights into the dynamics of the game.
\end{IEEEbiography}
\vspace{-0.6cm}
\begin{IEEEbiography}[{\includegraphics[width=1in,height=1.25in,clip,keepaspectratio]{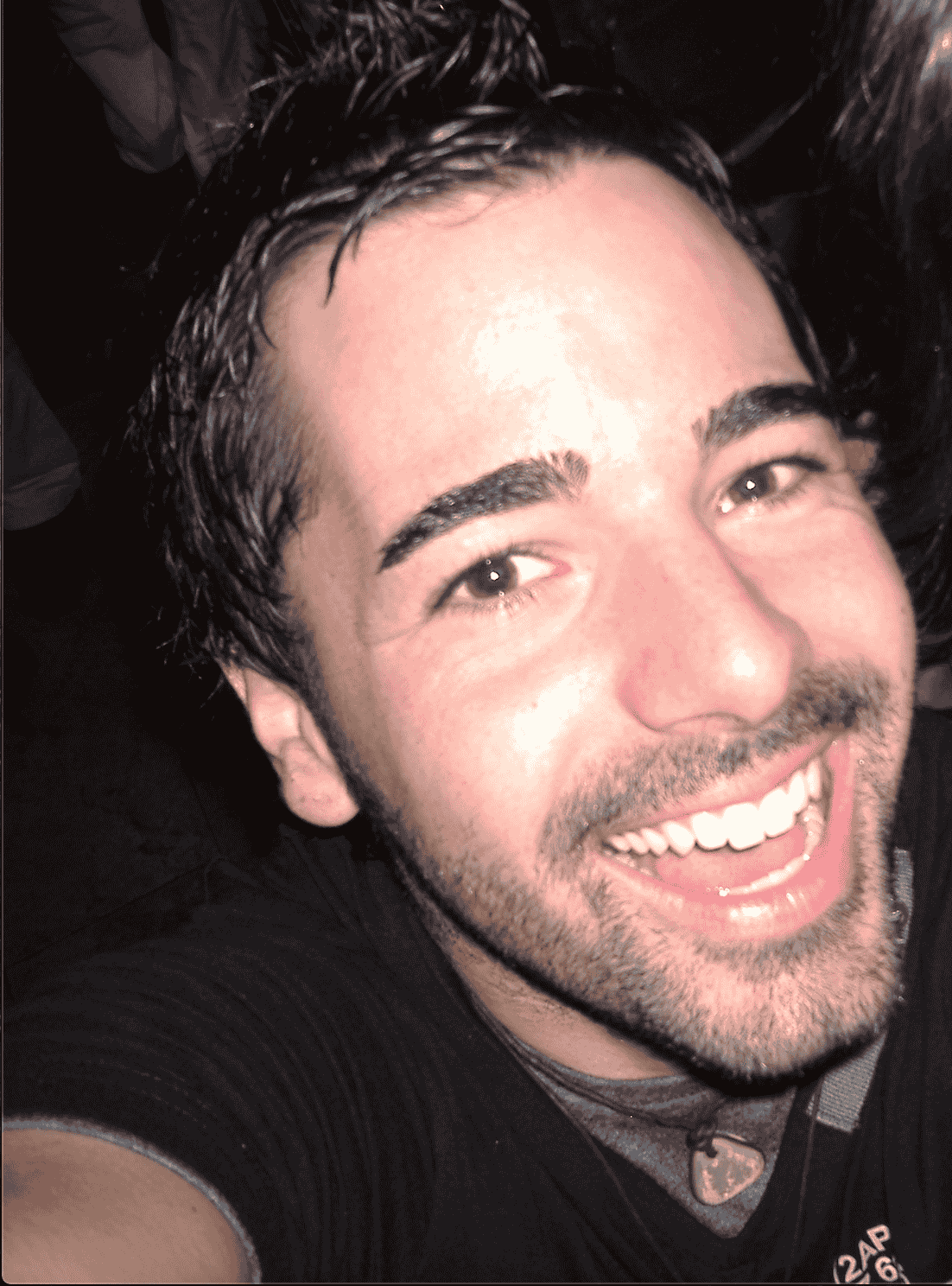}}]{Antonio Agudo}
received the M.Sc. degree in industrial engineering and electronics in 2010, M.Sc. degree in computer science in 2011, and the Ph.D. degree in computer vision and robotics in 2015, from University of Zaragoza. He was a visiting student with the vision group of Queen Mary University of London in 2013 and with the vision and imaging science group of University College London in 2014. He was also a visiting fellow at Harvard University in 2015, and at Universit\'e de Bordeaux in 2018 and 2019. He is a research scientist at Institut de Rob\`otica i Inform\`atica Industrial, CSIC-UPC; a recipient of the 2018 Best Paper Award Honorable Mention from the European Conference on Computer Vision, 2023 BBVA Leonardo Award and the 2024 Generative AI Art competition Winner from the International Conference on Image Processing.
\end{IEEEbiography}

\end{document}